\documentclass[journal]{IEEEtran}

\usepackage[mathscr]{eucal}
\usepackage[cmex10]{amsmath}
\usepackage{epsfig,epsf,psfrag}
\usepackage{amssymb,amsmath,amsthm,amsfonts,latexsym}
\usepackage{amsmath,graphicx,bm,xcolor,url}
\usepackage[caption=false]{subfig} 
\usepackage{fixltx2e}
\usepackage{array}
\usepackage{verbatim}
\usepackage{bm}
\usepackage{algorithmic, cite}
\usepackage{algorithm}
\usepackage{verbatim}
\usepackage{textcomp}
\usepackage{mathrsfs}
\usepackage{enumerate}
\usepackage{epstopdf}

\catcode`~=11 \def\UrlSpecials{\do\~{\kern -.15em\lower .7ex\hbox{~}\kern .04em}} \catcode`~=13 

\allowdisplaybreaks[3]

\newcommand{\calD}{\mathcal{D}}

\newcommand{\calG}{\mathcal{G}}

\newcommand{\calI}{\mathcal{I}}

\newcommand{\calL}{\mathcal{L}}
\newcommand{\calM}{\mathcal{M}}
\newcommand{\calN}{\mathcal{N}}
\newcommand{\calO}{\mathcal{O}}

\newcommand{\calR}{\mathcal{R}}
\newcommand{\calS}{\mathcal{S}}
\newcommand{\calT}{\mathcal{T}}

\newcommand{\ba}{\mathbf{a}}
\newcommand{\bA}{\mathbf{A}}

\newcommand{\bc}{\mathbf{c}}

\newcommand{\bd}{\mathbf{d}}
\newcommand{\bD}{\mathbf{D}}

\newcommand{\bE}{\mathbf{E}}

\newcommand{\bF}{\mathbf{F}}
\newcommand{\bg}{\mathbf{g}}
\newcommand{\bG}{\mathbf{G}}
\newcommand{\bh}{\mathbf{h}}

\newcommand{\bI}{\mathbf{I}}

\newcommand{\bP}{\mathbf{P}}
\newcommand{\bq}{\mathbf{q}}
\newcommand{\bQ}{\mathbf{Q}}

\newcommand{\bR}{\mathbf{R}}

\newcommand{\bt}{\mathbf{t}}
\newcommand{\bT}{\mathbf{T}}
\newcommand{\bu}{\mathbf{u}}
\newcommand{\bU}{\mathbf{U}}

\newcommand{\bV}{\mathbf{V}}

\newcommand{\bx}{\mathbf{x}}

\newcommand{\by}{\mathbf{y}}

\newcommand{\bz}{\mathbf{z}}

\DeclareMathAlphabet{\mathbsf}{OT1}{cmss}{bx}{n}
\DeclareMathAlphabet{\mathssf}{OT1}{cmss}{m}{sl}

\DeclareSymbolFont{bsfletters}{OT1}{cmss}{bx}{n}  
\DeclareSymbolFont{ssfletters}{OT1}{cmss}{m}{n}
\DeclareMathSymbol{\bsfGamma}{0}{bsfletters}{'000}
\DeclareMathSymbol{\ssfGamma}{0}{ssfletters}{'000}
\DeclareMathSymbol{\bsfDelta}{0}{bsfletters}{'001}
\DeclareMathSymbol{\ssfDelta}{0}{ssfletters}{'001}
\DeclareMathSymbol{\bsfTheta}{0}{bsfletters}{'002}
\DeclareMathSymbol{\ssfTheta}{0}{ssfletters}{'002}
\DeclareMathSymbol{\bsfLambda}{0}{bsfletters}{'003}
\DeclareMathSymbol{\ssfLambda}{0}{ssfletters}{'003}
\DeclareMathSymbol{\bsfXi}{0}{bsfletters}{'004}
\DeclareMathSymbol{\ssfXi}{0}{ssfletters}{'004}
\DeclareMathSymbol{\bsfPi}{0}{bsfletters}{'005}
\DeclareMathSymbol{\ssfPi}{0}{ssfletters}{'005}
\DeclareMathSymbol{\bsfSigma}{0}{bsfletters}{'006}
\DeclareMathSymbol{\ssfSigma}{0}{ssfletters}{'006}
\DeclareMathSymbol{\bsfUpsilon}{0}{bsfletters}{'007}
\DeclareMathSymbol{\ssfUpsilon}{0}{ssfletters}{'007}
\DeclareMathSymbol{\bsfPhi}{0}{bsfletters}{'010}
\DeclareMathSymbol{\ssfPhi}{0}{ssfletters}{'010}
\DeclareMathSymbol{\bsfPsi}{0}{bsfletters}{'011}
\DeclareMathSymbol{\ssfPsi}{0}{ssfletters}{'011}
\DeclareMathSymbol{\bsfOmega}{0}{bsfletters}{'012}
\DeclareMathSymbol{\ssfOmega}{0}{ssfletters}{'012}

\DeclareMathOperator*{\argmax}{arg\,max}

\DeclareMathOperator{\sgn}{sgn}

\DeclareMathOperator{\spn}{span}

\newtheorem{theorem}{Theorem} 
\newtheorem{lemma}[theorem]{Lemma}

\newtheorem{definition}{Definition}

\newtheorem{remark}{Remark}

\newtheorem{assumption}{Assumption}
\newtheorem{data model}{Data Model}

\newcommand{\qednew}{\nobreak \ifvmode \relax \else
      \ifdim\lastskip<1.5em \hskip-\lastskip
      \hskip1.5em plus0em minus0.5em \fi \nobreak
      \vrule height0.75em width0.5em depth0.25em\fi}

\begin{document}
\title{Innovation Pursuit: A New Approach to Subspace Clustering}

\author{Mostafa~Rahmani, \IEEEmembership{Student Member,~IEEE} and George~K.~Atia,~\IEEEmembership{Member,~IEEE} 
\thanks{This work was supported by NSF CAREER Award CCF-1552497 and NSF Grant CCF-1320547.

The authors are with the Department of Electrical and Computer Engineering, University of Central Florida, Orlando, FL 32816 USA (e-mails: mostafa@knights.ucf.edu, george.atia@ucf.edu).}
}

\markboth{}%
{Shell \MakeLowercase{\textit{et al.}}: Bare Demo of IEEEtran.cls for Journals}
\maketitle

\begin{abstract}
In subspace clustering, a group of data points belonging to a union of subspaces are assigned membership to their respective subspaces.
This paper presents a new approach dubbed Innovation Pursuit (iPursuit) to the problem of subspace clustering
%
using a new geometrical idea whereby subspaces are identified based on their relative novelties.
We present two frameworks in which the idea of innovation pursuit is used to distinguish the subspaces. Underlying the first framework is an iterative method that finds the subspaces consecutively by solving a series of simple linear optimization problems, each searching for a  direction of innovation in the span of the data potentially orthogonal to all subspaces except for the one to be identified in one step of the algorithm. A detailed mathematical analysis is provided establishing sufficient conditions for the proposed approach to correctly cluster the data points. The proposed approach can provably yield exact clustering even when the subspaces have significant intersections under mild conditions on the distribution of the data points in the subspaces. It is shown that the complexity of the iterative approach scales only linearly in the number of data points and subspaces, and quadratically in the dimension of the subspaces.
%
The second framework integrates iPursuit with spectral clustering to yield a new variant of spectral-clustering-based algorithms. The numerical simulations with both real and synthetic data demonstrate that iPursuit can often outperform the state-of-the-art subspace clustering algorithms, more so for subspaces with significant intersections, and that it significantly improves the state-of-the-art result for subspace-segmentation-based face clustering. 
\end{abstract}

\begin{IEEEkeywords}
Subspace Learning, Subspace Clustering, Linear Programming, Big Data, Innovation Pursuit, Unsupervised Learning.
\end{IEEEkeywords}

\IEEEpeerreviewmaketitle

\section{Introduction}
The grand challenge of contemporary data analytics and machine learning lies in dealing with ever-increasing amounts of high-dimensional data from multiple sources and different modalities.
The high-dimensionality of data increases the computational complexity and memory requirements of existing algorithms and can adversely degrade their performance \cite{slavakis2014modeling,7968311}.
However, the observation that high-dimensional datasets often have intrinsic low-dimensional structures has enabled some noteworthy progress in analyzing such data.
For instance, the high-dimensional digital facial images under different illumination were shown to approximately lie in a very low-dimensional subspace, which led to the development of efficient algorithms that leverage low-dimensional representations of such images~\cite{basri2003lambertian ,wright2009robust}.

Linear subspace models are widely used in   data analysis since many datasets can be well-approximated with low-dimensional subspaces~\cite{bishop2006pattern}.
%
When data in a high-dimensional space lies in a single subspace, conventional techniques such as Principal Component Analysis  can be efficiently used to find the underlying low-dimensional subspace~\cite{rahmani2015randomized,zhang2014novel,lerman2015robust}. However, in many applications the data points may be originating from multiple independent sources, in which case a union of subspaces can better model the data~\cite{vidal2011subspace}.

Subspace clustering is concerned with learning these low-dimensional subspaces and clustering the data points to their respective subspaces \cite{elhamifar2013sparse,heckel2013robust,rahmani2017innovation,liu2013robust,soltanolkotabi2012geometric,vidal2011subspace,vidal2005generalized,zhang2009median,chen2009spectral}. This problem arises in many applications, including computer vision (e.g. motion segmentation \cite{vidal2011subspace}, face clustering \cite{ho2003clustering}), gene expression analysis \cite{mcwilliams2014subspace,kriegel2009clustering}, and image processing \cite{yang2008unsupervised}. Some of the difficulties associated with subspace clustering are that neither the number of subspaces nor their dimensions are known, in addition to the unknown membership of the data points to the subspaces.
%

\subsection{Related work}
Numerous approaches for subspace clustering have been studied in prior work, including statistical-based approaches \cite{yang2006robust}, spectral clustering \cite{elhamifar2013sparse}, the algebraic-geometric approach \cite{vidal2005generalized} and iterative methods \cite{bradley2000k}.  In this section, we briefly discuss some of the most popular existing approaches for subspace clustering. We refer the reader to \cite{vidal2011subspace} for a comprehensive survey on the topic.
Iterative algorithms such as \cite{bradley2000k,zhang2009median} were some of the first methods addressing the multi-subspace learning problem. These algorithms alternate between assigning the data points to the identified subspaces and updating the subspaces. Some of the drawbacks of this class of algorithms is that they can converge to a local minimum and typically assume that the dimension of the subspaces and their number are known.

Another reputable idea for subspace segmentation is based on the algebraic geometric approach. These algorithms, such as the Generalized Principal Component Analysis (GPCA) \cite{vidal2005generalized}, fit the data using a set of polynomials whose gradients at a point are orthogonal to the subspace containing that point. GPCA does not impose any restrictive conditions on the subspaces (they do not need to be independent), 
albeit it is sensitive to noise and has exponential complexity in the number of subspaces and their dimensions.

A class of clustering algorithms, termed statistical clustering methods, make some assumptions about the distribution of the data in the subspaces. For example, the iterative algorithm in~\cite{stat1,stat2} assumes that the distribution of the data points in the subspaces is Gaussian. These algorithms typically require prior specifications for the number and dimensions of the subspaces, and are generally sensitive to initialization. Random sample consensus (RANSAC) is another iterative statistical method for robust model fitting \cite{rnc1}, which recovers one subspace at a time by repeatedly sampling small subsets of data points 
and identifying a consensus set consisting of all the points in the entire dataset that belong to the subspace spanned by the selected points. The consensus set is removed and the steps are repeated until all the subspaces are identified. The main drawback of this approach is scalability since  the number of trials required to select points in the same subspace grows exponentially with the number and dimension of the subspaces.

Much of the recent research work on subspace clustering is focused on spectral clustering \cite{von2007tutorial} based methods \cite{dyer2013greedy,gao2015multi,elhamifar2013sparse,heckel2013robust,liu2013robust,rahmani2017direction,soltanolkotabi2012geometric,wang2013provable,chen2009spectral,park2014greedy}. These algorithms consist of two main steps and mostly differ in the first step. In the first step, a similarity matrix is constructed by finding a neighborhood for each data point, and in the second step spectral clustering \cite{von2007tutorial} is applied to the similarity matrix. 
Recently, several spectral clustering based algorithms were proposed with both
theoretical guarantees and superior empirical performance. The  Sparse Subspace Clustering (SSC)~\cite{elhamifar2013sparse} uses $\ell_1$-minimization for neighborhood construction. In \cite{soltanolkotabi2012geometric}, it was shown that under certain conditions, SSC can yield exact clustering even for subspaces with intersection. Another algorithm called Low-Rank Representation (LRR) \cite{liu2013robust} uses nuclear norm minimization to find the neighborhoods (i.e., build the similarity matrix). LRR is robust to outliers but has provable guarantees only when the data is drawn from independent subspaces.
In this paper, we show that the idea of innovation pursuit underlying the proposed iterative subspace clustering method can also be used to design a new spectral clustering based method. The result spectral clustering based algorithm is shown to notably outperform the existing spectral clustering based methods.

\subsection{Contributions}

The proposed method advances the state-of-the-art research in subspace clustering on several fronts. First, iPursuit rests on a novel geometrical idea whereby the subspaces are identified by searching the directions of innovation in the span of the data.
Second, to the best of our knowledge this is the first scalable \emph{iterative algorithm} with provable guarantees -- the computational complexity of iPursuit only scales linearly in the number of subspaces and quadratically in their dimensions (c.f. Section \ref{sec:comp_analysis}). By contrast, GPCA \cite{vidal2005generalized,vidal2011subspace} (without spectral clustering) and RANSAC \cite{rnc1}, which are popular iterative algorithms, have exponential complexity in the number of subspaces and their dimensions. Third, innovation pursuit in the data span enables superior performance when the subspaces have considerable intersections, and brings about substantial speedups, in comparison to the existing subspace clustering approaches. Fourth, the formulation enables many variants of the algorithm to inherit robustness properties in highly noisy settings (c.f. Section \ref{sec:noisy}).
\textcolor{black}{Fifth, the proposed innovation search approach can be integrated with spectral clustering to yield a new spectral clustering based algorithm, which is shown to notably outperform the existing spectral clustering based methods and yields the state-of-the-art results in the challenging face clustering problem. }

\subsection{Notation and definitions}
Bold-face upper-case letters are used to denote matrices and bold-face lower-case letters are used to denote vectors. Given a matrix $\bA$, $\| \bA \|$ denotes its spectral norm. For a vector $\ba$, $\| \ba \|$ denotes its $\ell_2$-norm and $\| \ba \|_1$ its $\ell_1$-norm. Given two matrices $\bA_1$ and $\bA_2$ with equal number of rows, the matrix
$
\bA_3 = [\bA_1 \: \: \bA_2]
$
is the matrix formed from the concatenation of $\bA_1$ and $\bA_2$. Given matrices $\{ \bA_i \}_{i=1}^n$ with equal number of rows, we use the union symbol $\cup$ to define the matrix
\[
\overset{n} {\underset{i = 1 }{\cup}} \bA_i =  [ \bA_1 \: \:  \bA_2 \:  \: ... \: \:  \bA_n ]
\]
as the concatenation of the matrices $\{ \bA_i \}_{i=1}^n$. For a matrix $\bD$, we overload the set membership operator by using the notation $\bd\in \bD$ to signify that $\bd$ is a column of $\bD$. A collection of subspaces $\{ \calG_i \}_{i=1}^n$ is said to be independent if
$
\operatorname{dim} \left(\overset{n} {\underset{i = 1 }{\oplus}} \calG_i \right) = \sum_{i=1}^n \operatorname{dim} (\calG_i)
$
where $\oplus$ denotes the direct sum operator and $\operatorname{dim} (\calG_i)$ is the dimension of $\calG_i$. Given a vector $\ba$, $\big | \ba | $ is the vector whose elements are equal to the absolute value of the elements of $\ba$. For a real number $a$, $\sgn(a)$ is equal to $1$ if $a > 0$, $-1$ if $a < 0$, and $0$ if $a = 0$.
 The complement of a set $\cal L$ is denoted ${\cal L}^c$. Also, for any positive integer $n$, the index set $\{1,\ldots, n\}$ is denoted $[n]$.

\subsubsection*{Innovation subspace}
Consider two subspaces $\mathcal{S}_1$ and $\mathcal{S}_2$, such that $\mathcal{S}_1\ne \mathcal{S}_2$, and one is not contained in the other. 
This means that each subspace carries some innovation w.r.t. the other.
As such, corresponding to each subspace we define an innovation subspace, which is its novelty (innovation) relative to the other subspaces. More formally, the innovation subspace is defined as follows.

\begin{definition}
Assume that $\bV_1$ and $\bV_2$ are two orthonormal bases for $\mathcal{S}_1$ and $\mathcal{S}_2$, respectively. We define the subspace $\mathcal{I} \left( \mathcal{S}_2 \perp \mathcal{S}_1 \right)$ as the innovation subspace of $\mathcal{S}_2$ over $\mathcal{S}_1$ that is spanned by
$
\left( \bI - \bV_1 \bV_1^T \right) \bV_2.
$
In other words, $\mathcal{I} \left( \mathcal{S}_2 \perp \mathcal{S}_1 \right)$ is the complement of $\mathcal{S}_1$ in the subspace $\mathcal{S}_1 \oplus \mathcal{S}_2$.
\label{def1}
\end{definition}

\begin{figure}[t!]
 \centering
    \includegraphics[width=0.30\textwidth]{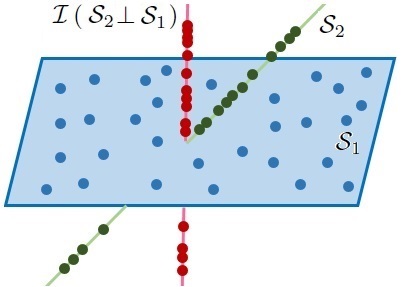}
    \vspace{-0.3cm}
    \caption{The subspace $\mathcal{I} \left( \mathcal{S}_2 \perp \mathcal{S}_1 \right)$ is the innovation subspace of $\calS_2$ over $\calS_1$. The subspace $\mathcal{I} \left( \mathcal{S}_2 \perp \mathcal{S}_1 \right)$ is orthogonal to $\calS_1$. }
    \label{innovationfig}
\end{figure}

In a similar way, we can define $\mathcal{I} \left( \mathcal{S}_1 \perp \mathcal{S}_2 \right)$ as the innovation subspace of $\mathcal{S}_1$ over $\mathcal{S}_2$. The subspace $\mathcal{I} \left( \mathcal{S}_1 \perp \mathcal{S}_2 \right)$ is the complement of $\mathcal{S}_2$ in $\mathcal{S}_1 \oplus \mathcal{S}_2$. Fig. \ref{innovationfig} illustrates a scenario in which the data lies in a two-dimensional subspace $\calS_1$, and a one-dimensional subspace $\calS_2$. The innovation subspace of $\calS_2$ over $\calS_1$ is orthogonal to $\calS_1$. Since $\calS_1$ and $\calS_2$ are independent, $\calS_2$ and $\calI \left( \calS_2 \perp \calS_1 \right)$ have equal dimension. It is easy to see that the dimension of $\mathcal{I} \left( \mathcal{S}_2 \perp \mathcal{S}_1 \right)$ is equal to the dimension of $\mathcal{S}_2$ minus the dimension of $\mathcal{S}_1 \cap \mathcal{S}_2$.

\section{Proposed Approach}
\label{sec:proposed_secd}
In this section, 
the core idea underlying iPursuit is described by first introducing a non-convex optimization problem. Then, we propose a convex relaxation and show that solving the convex problem yields the correct subspaces under mild sufficient conditions. 
First, we present the idea of iPursuit and its analysis within the proposed iterative framework. In Section \ref{sec:integr}, we present an alternative framework wherein the proposed innovation search approach is integrated with spectral clustering to yield a new variant of spectral-clustering-based algorithms. 

In this section, it is assumed that the given data matrix follows the following data model. 
\\
\\
\textbf{Data Model 1.} 
The data matrix $\bD \in \mathbb{R}^{M_1 \times M_2}$ can be represented as
$
\bD = [\bD_1 \: ... \: \bD_N] \bT
$
where $\bT$ is an arbitrary permutation matrix. The columns of $\bD_i \in \mathbb{R}^{M_1 \times n_i}$ lie in $\mathcal{S}_i$, where $\mathcal{S}_i$ is an $r_i$-dimensional linear subspace, for $1 \leq i \leq N$, and, $\sum_{i = 1}^N n_i = M_2$. Define $\bV_i$ as an orthonormal basis for $\mathcal{S}_i$. In addition, define $\mathcal{D}$ as the space spanned by the data, i.e., $ \mathcal{D} =  \overset{N} {\underset{i = 1 }{\oplus}} \mathcal{S}_i$. 
 Moreover, it is assumed that any subspace in the set of subspaces $\{ \calS_i \}_{i=1}^N$ has an innovation over the other subspaces, to say that, for $1 \leq i \leq N$, the subspace $\calS_i$ does not completely lie in $\overset{N} {\underset{k = 1 \atop k \neq i }{\oplus}} \mathcal{S}_k \:$. In addition, the columns of $\bD$ are normalized, i.e., have unit $\ell_2$-norm. 





\subsection{Innovation pursuit: Insight}
\label{sec:idea_iPursuit}
iPursuit is a multi-step algorithm that identifies one subspace at a time. In each step, the data is clustered into two subspaces. One subspace is the identified subspace and the other one is the direct sum of the other subspaces. The data points of the identified subspace are removed and the algorithm is applied to the remaining data to find the next subspace.
Accordingly, each step of the algorithm can be interpreted as a subspace clustering problem with two subspaces.
Therefore, for ease of exposition we first investigate the two-subspace scenario then extend the result to multiple (more than two) subspaces. Thus, in this subsection, it is assumed that the data follows Data model 1 with $N = 2$.

To gain some intuition, we consider an example before stating our main result. Consider the case where $\mathcal{S}_1$ and $\mathcal{S}_2$ are not orthogonal and assume that $n_2 < n_1$. The non-orthogonality of $\calS_1$ and $\calS_2$ is not a requirement, but is merely used herein to easily explain the idea underlying the proposed approach. Let $\bc^{*}$ be the optimal point of the following optimization problem
\begin{eqnarray}
\begin{aligned}
 \underset{\hat{\bc}}{\min}
 \| \hat{\bc}^T \bD \|_0
 \quad \text{s. t.} \quad
 \hat{\bc} \in \mathcal{D} \quad \text{and} \quad  \| \hat{\bc} \| = 1 ,
\end{aligned}
\label{non-convex}
\end{eqnarray}
where $\| . \|_0$ is the $\ell_0$-norm. Hence, $\| \hat{\bc}^T \bD \|_0$ is equal to the number of non-zero elements of $\hat{\bc}^T \bD $. The first constraint forces the search for $\hat{\bc}$ in the span of the data, and the equality constraint $\| \hat{\bc} \| = 1$ is used to avoid the trivial $\hat{\bc} = 0$ solution. Assume that the data points are distributed in $\mathcal{S}_1$ and $\mathcal{S}_2$ uniformly at random. Thus, the data is not aligned with any direction in $\mathcal{S}_1$ and $\mathcal{S}_2$ with high probability (whp).

The optimization problem (\ref{non-convex}) searches for a non-zero vector in the span of the data that is orthogonal to the maximum number of data points. We claim that the optimal point of (\ref{non-convex}) lies in $\mathcal{I} \left( \mathcal{S}_2 \perp \mathcal{S}_1 \right)$ whp given the assumption that the number of data points in $\mathcal{S}_1$ is greater than the number of data points in $\mathcal{S}_2$. In addition, since the feasible set of (\ref{non-convex}) is restricted to $\mathcal{D}$, there is no feasible vector that is orthogonal to the whole data.  To further clarify this argument, consider the following scenarios:
\begin{enumerate}[I.]
\item If $\bc^{*}$ lies in $\mathcal{S}_1$, then it cannot be orthogonal to most of the data points in $\mathcal{S}_1$ since the data is uniformly distributed in the subspaces. In addition, it cannot be orthogonal to the data points in $\mathcal{S}_2$ given that $\calS_1$ and $\calS_2$ are not orthogonal. Therefore, the optimal point of ($\ref{non-convex}$) cannot be in $\mathcal{S}_1$ given that the optimal vector should be orthogonal to the maximum number of data points. Similarly, the optimal point cannot lie in $\mathcal{S}_2$.

\item If $\bc^{*}$ lies in $\mathcal{I} \left( \mathcal{S}_1 \perp \mathcal{S}_2 \right)$, then it is orthogonal to the data points in $\bD_2$. However, $n_2 < n_1$. Thus, If $\bc^{*}$ lies in $\mathcal{I} \left( \mathcal{S}_2 \perp \mathcal{S}_1 \right)$ (which is orthogonal to $\mathcal{S}_1$) the cost function of ($\ref{non-convex}$) can be decreased.

\item If $\bc^{*}$ lies in none of the subspaces $\mathcal{S}_1$, $\mathcal{S}_2$, $\mathcal{I} \left( \mathcal{S}_2 \perp \mathcal{S}_1 \right)$ and $\mathcal{I} \left( \mathcal{S}_2 \perp \mathcal{S}_1 \right)$, then it is not orthogonal to $\mathcal{S}_1$ nor $\mathcal{S}_2$. Therefore, $\bc^{*}$ cannot be orthogonal to the maximum number of data points.
\end{enumerate}
Therefore, the algorithm is likely to choose the optimal point from $\mathcal{I} \left( \mathcal{S}_2 \perp \mathcal{S}_1 \right)$. Thus, if $\bc^{*} \in \mathcal{I} \left( \mathcal{S}_2 \perp \mathcal{S}_1 \right)$, we can obtain $\calS_2$ from the span of the columns of $\bD$ corresponding to the non-zero elements of $(\bc^{*})^T \bD$. The following lemma ensures that these columns span $\calS_2$.

\begin{lemma}
The  columns of $\bD$ corresponding to the non-zero elements of $(\bc^{*})^T \bD$ span $\calS_2$ if both conditions below are satisfied:
\begin{enumerate}[i)]
\item $\bc^{*} \in \mathcal{I} \left( \mathcal{S}_2 \perp \mathcal{S}_1 \right)$.
\item $\bD_2$ cannot follow Data model 1 with $N > 1$, that is, the data points in $\bD_2$ do not lie in the union of lower dimensional subspaces within $\mathcal{S}_2$ each with innovation w.r.t. to the other subspaces.
\end{enumerate}
\label{new_lemma}
\end{lemma}
%


\noindent
\textcolor{black}{
We note that if the second requirement of Lemma \ref{new_lemma} is not satisfied, the columns of $\bD_2$ follow Data model 1 with $N \ge 2$, in which case the problem can be viewed as one of subspace clustering with more than two subspaces. 
The clustering problem with more than two subspaces will be addressed in Section \ref{sec:multiple}. 
}

\begin{remark}
At a high level, the innovation search optimization problem (\ref{non-convex}) finds the most sparse vector in the row space of $\bD$. Interestingly, finding the most sparse vector in a linear subspace has bearing on, and has been effectively used in, other machine learning problems, including dictionary learning and spectral estimation \cite{qu2014finding,spielman2013exact}. 
\textcolor{black}{
In addition, it is interesting to note that in contrast to SSC which finds the most sparse vectors in the null space of the data, iPursuit searches for the most sparse vector in the row space of the data.} 
\end{remark}

\begin{figure}[t!]
	\centering
    \includegraphics[width=0.45\textwidth]{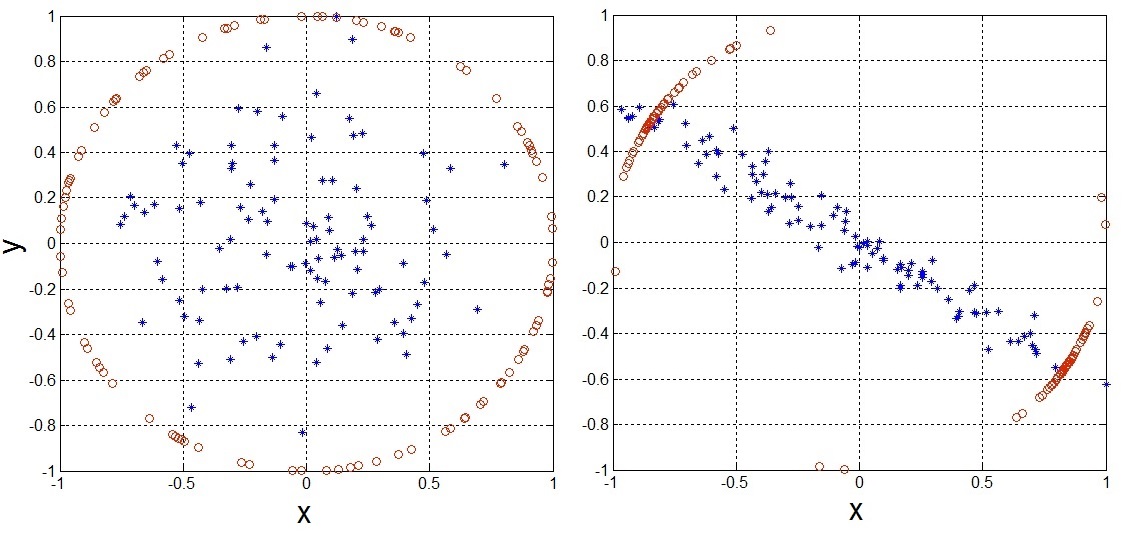}
    \vspace{-.1in}
    \caption{ Data distributions in a two-dimensional subspace. The blue stars and red circles are the data points and their  projections on the unit circle, respectively. In the left plot, the data points are distributed uniformly at random. Thus, they are not aligned along any specific directions and the permeance statistic cannot be small. In the right plot, the data points are aligned, hence the permeance statistic is small.}
    \label{data_distri}
\end{figure}

\subsection{Convex relaxation}
\label{sec:relarex_sec}
The cost function of (\ref{non-convex}) is non-convex and the combinatorial $\ell_0$-norm minimization may not be computationally tractable. Since the $\ell_1$-norm is known to provide an efficient convex approximation of the $\ell_0$-norm, we relax the non-convex cost function and rewrite (\ref{non-convex}) as
\begin{eqnarray}
\begin{aligned}
 \underset{\hat{\bc}}{\min}
 \| \hat{\bc}^T \bD \|_1
  \quad \text{s. t.} \quad
 \hat{\bc} \in \mathcal{D} \quad \text{and} \quad  \| \hat{\bc} \| = 1.
\end{aligned}
\label{nime_convex}
\end{eqnarray}
The optimization problem (\ref{nime_convex}) is still non-convex in view of the non-convexity of its feasible set. Therefore, we further substitute the equality constraint with a linear constraint and rewrite (\ref{nime_convex}) as
\begin{eqnarray}
(\text{IP})\hspace{.5cm}\begin{aligned}
 \underset{\hat{\bc}}{\min}
 \| \hat{\bc}^T \bD \|_1 \quad \text{s. t.}
\quad \hat{\bc} \in \mathcal{D} \quad \text{and} \quad   \hat{\bc}^T \bq = 1.
\end{aligned}
\label{convex}
\end{eqnarray}
(IP) is the core program of iPursuit to find a direction of innovation. Here, $\bq$ is a unit $\ell_2$-norm vector which is not orthogonal to $\mathcal{D}$. The vector $\bq$ can be chosen as a random unit vector in $\calD$. In Section \ref{sec:choose_q}, we develop a methodology to learn a good choice for $\bq$ from the given data matrix. The relaxation of the quadratic equality constraint to a linear constraint is a common technique in the literature \cite{spielman2013exact}.

\subsection{Segmentation of two subspaces: Performance guarantees}
\label{sec:2sub_perf}
Based on Lemma \ref{new_lemma}, to show that the proposed program (\ref{convex}) yields correct clustering, it suffices to show that the optimal point of (\ref{convex}) lies in $\mathcal{I} \left( \mathcal{S}_2 \perp  \mathcal{S}_1 \right)$ given that condition ii of Lemma 1 is satisfied, or lies in $\mathcal{I} \left( \mathcal{S}_1 \perp  \mathcal{S}_2 \right)$ given that condition ii of Lemma 1 is satisfied for $\bD_1$. The following theorem provides sufficient conditions for the optimal point of (\ref{convex}) to lie in $\mathcal{I} \left( \mathcal{S}_2 \perp  \mathcal{S}_1 \right)$ provided that
\begin{eqnarray}
\underset{\bc \in \mathcal{I} \left( \mathcal{S}_2 \perp  \mathcal{S}_1 \right) \atop  \bc^T \bq = 1}{\inf} \: \:  \| \bc^T \bD \|_1 < \underset{\bc \in \mathcal{I}  \left( \mathcal{S}_1 \perp  \mathcal{S}_2 \right) \atop  \bc^T \bq = 1}{\inf} \: \:  \| \bc^T \bD \|_1.
\label{cond1}
\end{eqnarray}
If the inequality in (\ref{cond1}) is reversed, then parallel sufficient conditions can be established for the optimal point of (\ref{convex}) to lie in the alternative subspace $\mathcal{I} \left( \mathcal{S}_1 \perp  \mathcal{S}_2 \right)$. Hence, assumption (\ref{cond1}) does not lead to any loss of generality.
%

Since $\mathcal{I} \left( \mathcal{S}_2 \perp  \mathcal{S}_1 \right)$ and $\mathcal{I} \left( \mathcal{S}_1 \perp  \mathcal{S}_2 \right)$ are orthogonal to $\mathcal{S}_1$ and $\mathcal{S}_2$, respectively, condition (\ref{cond1}) is equivalent to
\begin{eqnarray}
\underset{\bc \in \mathcal{I} \left( \mathcal{S}_2 \perp  \mathcal{S}_1 \right) \atop   \bc^T \bq = 1}{\inf} \: \:  \| \bc^T \bD_2 \|_1 < \underset{\bc \in \mathcal{I}  \left( \mathcal{S}_1 \perp  \mathcal{S}_2 \right) \atop  \bc^T \bq = 1}{\inf} \: \:  \| \bc^T \bD_1 \|_1.
\label{cond_essential}
\end{eqnarray}
Conceptually, assumption (\ref{cond_essential}) is related to the assumption $n_2 < n_1$ used in the example of Section \ref{sec:idea_iPursuit} in the sense that it makes it more likely for the direction of innovation to lie in $\mathcal{I} \left( \mathcal{S}_2 \perp  \mathcal{S}_1 \right)$\footnote{Henceforth, when the two-subspace scenario is considered and (\ref{cond_essential}) is satisfied, the innovation subspace refers to $\mathcal{I} \left( \mathcal{S}_2 \perp  \mathcal{S}_1 \right).$}.

The sufficient conditions of Theorem \ref{two_ind} for the optimal point of (\ref{convex}) to lie in $\mathcal{I} \left( \mathcal{S}_2 \perp  \mathcal{S}_1 \right)$ are characterized in terms of the optimal solution to an oracle optimization problem (OP), where the feasible set of (IP) is replaced by $\mathcal{I} \left( \mathcal{S}_2 \perp  \mathcal{S}_1 \right)$. The oracle problem (OP) is defined as
%
\begin{eqnarray}
\label{eq:oracel}
(\text{OP})\hspace{.5cm}\begin{aligned}
& \underset{\hat{\bc}}{\min}
& & \| \hat{\bc}^T \bD_2 \|_1 \\
& \text{subject to}
& & \hat{\bc} \in \mathcal{I} \left( \mathcal{S}_2 \perp  \mathcal{S}_1 \right) \quad \text{and} \quad   \hat{\bc}^T \bq = 1. \\
\end{aligned}
\end{eqnarray}
Before we state the theorem, we also define the index set ${\cal L}_0$ comprising the indices of the columns of $\bD_2$ orthogonal to $\bc_2$ (the optimal solution to (OP)),
\begin{align}
{\cal L}_0 = \{i \in [n_2] : \bc_2^T \bd_i = 0, \bd_i \in \bD_2\},
\label{eq:set_L_0}
\end{align}
with cardinality $n_0 = |{\cal L}_0|$ and a complement set ${\cal L}_0^c$.

\begin{theorem}
Suppose the data matrix $\bD$ follows Data model 1 with $N=2$. Also, assume that condition (\ref{cond_essential}) and the requirement of Lemma \ref{new_lemma} for $\bD_2$ are satisfied (condition ii of Lemma \ref{new_lemma}). Let $\bc_2$ be the optimal point of the oracle program (OP) and
define
\begin{eqnarray}
\alpha = \sum_{\bd_i \in \bD_2 } \sgn (\bc_2^T \bd_i) \bd_i
\label{eq:alpha_def}
\end{eqnarray}
Also, let $\bP_2$ denote an orthonormal basis for $\mathcal{I} \left( \mathcal{S}_2 \perp  \mathcal{S}_1 \right)$, $n_0$ the cardinality of ${\cal L}_0$ defined in (\ref{eq:set_L_0}), and assume that $\bq$ is a unit $\ell_2$-norm vector in $\calD$ that is not orthogonal to $\mathcal{I} \left( \mathcal{S}_2 \perp  \mathcal{S}_1 \right)$.
If
\begin{eqnarray}
\begin{aligned}
& \frac{1}{2} \underset{\delta \in \mathcal{S}_1  \atop \| \delta \| = 1}{\inf} \sum_{\bd_i \in \bD_1} \left| \mathbf{\delta}^T \bd_i \right| >    \| \bV_1^T \bV_2 \|  \bigg( \|\alpha\| + n_0 \bigg) \: , \: \:  \text{and} \\
& \frac{ \| \bq^T \bP_2 \|}{ 2\| \bq^T \bV_1 \|}  \left( \underset{\delta \in \mathcal{S}_1  \atop \| \delta \| = 1}{\inf} \sum_{\bd_i \in \bD_1} \left| \mathbf{\delta}^T \bd_i \right|  \right) >   \| \bV_2^T \bP_2 \| \bigg( \| \alpha \| + n_0 \bigg),
\end{aligned}
\label{lemma_inq}
\end{eqnarray}
then $\bc_2$, which lies in $\mathcal{I} \left( \mathcal{S}_2 \perp  \mathcal{S}_1 \right)$, is the optimal point of (IP) in (\ref{convex}), and iPursuit clusters the data correctly.
\label{two_ind}
\end{theorem}

In what follows, we provide a detailed discussion of
the significance of the sufficient conditions (\ref{lemma_inq}), which
reveal some interesting facts about the properties of iPursuit. 


\smallbreak
\noindent 1. \textit{The distribution of the data matters.}\\
The LHS of (\ref{lemma_inq}) is known as the permeance statistic \cite{lerman2015robust}. For a set of data points $\bD_i$ in a subspace $\mathcal{S}_i$, the permeance statistic is defined as
\begin{eqnarray}
\begin{aligned}
\mathcal{P}(\bD_i , \mathcal{S}_i)= \underset{\bu \in \mathcal{S}_i  \atop \| \bu \| = 1}{\inf} \sum_{\bd_i \in \bD_i} \left| \mathbf{u}^T \bd_i \right|.
\end{aligned}
\end{eqnarray}
The permeance statistic is an efficient measure of how well the data is distributed in the subspace. 
Fig. \ref{data_distri} illustrates two scenarios for the distribution of data in a two-dimensional subspace. In the left plot, the data points are distributed uniformly at random. In this case, the permeance statistic cannot be small since the data points are not concentrated along any directions. In the right plot, the data points are concentrated along some direction and hence the data is not well distributed in the subspace. In this case, we can find a direction along which the data has small projection.  

Having $n_0$ on the RHS underscores the relevance of the distribution of the data points within $\calS_2$ since $\bc_2$ cannot be simultaneously orthogonal to a large number of columns of $\bD_2$ if the data does not align along particular directions. Hence, the distribution of the data points within each subspace has bearing on the performance of iPursuit. \textcolor{black}{We emphasize that the uniform distribution of the data points is not a requirement of the algorithm as shown in the numerical experiments. Rather, it is used in the proof of Theorem \ref{two_ind}, which establishes sufficient conditions for correct subspace identification under uniform data distribution in worst case scenarios. 
}

\smallbreak
\noindent 2. \textit{The coherency of $\bq$ with $\mathcal{I} \left( \mathcal{S}_2 \perp \mathcal{S}_1 \right)$ is an important factor.}\\
An important performance factor in iPursuit is the coherency of the vector $\bq$ with the subspace  $\mathcal{I} \left( \mathcal{S}_2 \perp \mathcal{S}_1 \right)$. To clarify, suppose that (\ref{cond_essential}) is satisfied and assume that the vector $\bq$ lies in  $\mathcal{D}$. If the optimal point of (\ref{convex}) lies in $\mathcal{I} \left( \mathcal{S}_2 \perp \mathcal{S}_1 \right)$, iPursuit will yield exact clustering. However, if $\bq$ is strongly coherent with $\mathcal{S}_1$ (i.e., the vector $\bq$ has small projection on $\mathcal{I} \left( \mathcal{S}_2 \perp \mathcal{S}_1 \right)$), then the optimal point of (\ref{convex}) may not lie in $\mathcal{I} \left( \mathcal{S}_2 \perp \mathcal{S}_1 \right)$. The rationale is that the Euclidean norm of any feasible point of (\ref{convex}) lying in $\mathcal{I} \left( \mathcal{S}_2 \perp \mathcal{S}_1 \right)$ will have to be large to satisfy the equality constraint when $\bq$ is incoherent with $\mathcal{I} \left( \mathcal{S}_2 \perp \mathcal{S}_1 \right)$, which in turn would increase the cost function.
As a matter of fact, the factor
$
\frac{ \| \bq^T \bP_2 \|}{ \| \bq^T \bV_1 \|}
$
in the second inequality of (\ref{lemma_inq})  confirms our intuition about the importance of the coherency of $\bq$ with $\mathcal{I} \left( \mathcal{S}_2 \perp \mathcal{S}_1 \right)$. In particular, (\ref{lemma_inq})  suggests that iPursuit could have more difficulty yielding correct clustering if the projection of $\bq$ on the subspace $\mathcal{S}_1$ is increased (i.e., the projection of the vector on the subspace $\mathcal{I} \left( \mathcal{S}_2 \perp \mathcal{S}_1 \right)$ is decreased).
The coherence property could have a more serious effect on the performance of the algorithm for non-independent subspaces, especially when the dimension of their intersection is significant. For instance, consider the scenario where the vector $\bq$ is chosen randomly from $\calD$, and define $y$ as the dimension of the intersection of $\calS_1$ and $\calS_2$. It follows that $\mathcal{I} \left( \mathcal{S}_2 \perp \mathcal{S}_1 \right)$ has dimension $r_2 - y$. Thus,
$
 \frac{ \mathbb{E} \left\{ \| \bq^T \bP_2 \|  \right\} }{ \mathbb{E} \left\{\| \bq^T \bV_1 \| \right\}} = \frac{r_2 - y}{r_1}.
$
Therefore, a randomly chosen vector $\bq$ is likely to have a small projection on the innovation subspace when $y$ is large. As such, in dealing with subspaces with significant intersection, it may not be favorable to choose the vector $\bq$ at random. In Section \ref{sec:choose_q} and section \ref{sec:noisy}, we develop a simple technique to learn a good choice for $\bq$ from the given data. This technique makes iPursuit remarkably powerful in dealing with subspaces with intersection as shown in the numerical results section.

Now, we demonstrate that the sufficient conditions (\ref{lemma_inq}) are not restrictive.
The following lemma simplifies the sufficient conditions of Theorem \ref{two_ind} when the data points are randomly distributed in the subspaces. In this setting, we show that the conditions are naturally satisfied.
\begin{lemma}
 \textcolor{black}{ Assume $\bD$ follows Data model 1 with $N=2$ and the data
points are drawn uniformly at random from the intersection of the unit sphere $\mathbb{S}^{M_1 -1 }$ and each subspace. If}
%
\begin{eqnarray}
\begin{aligned}
&\sqrt{\frac{2}{\pi}} \frac{n_1}{r_1} - 2\sqrt{n_1} -t_1 \sqrt{\frac{n_1}{r_1 -1 }}  \\
&\quad\quad\quad\quad\quad\quad\quad\quad > 2 \| \bV_1^T \bV_2 \| \bigg( t_2  \sqrt{n_2 - n_0} + n_0 \bigg), \\
\\
& \frac{ \| \bq^T \bP_2 \|}{ \| \bq^T \bV_1 \|}  \left( \sqrt{\frac{2}{\pi}} \frac{n_1}{r_1} - 2\sqrt{n_1} -t_1 \sqrt{\frac{n_1}{r_1 -1 }} \right)  \\
&\quad\quad\quad\quad\quad\quad\quad\quad > 2 \| \bV_2^T \bP_2 \| \bigg( t_2  \sqrt{n_2 - n_0} + n_0 \bigg),
\end{aligned}
\label{random_suf}
\end{eqnarray}
then the optimal point of (\ref{convex}) lies in $\mathcal{I} \left( \mathcal{S}_2 \perp  \mathcal{S}_1 \right)$ with probability at least
$
1 - \exp \left(-\frac{r_2}{2}(t_2^2 - \log (t_2^2) -1)  \right) - \exp \left( -\frac{t_1^2}{2} \right) ,
$
for all $t_2 > 1 \: , \: t_1 \ge 0$.
\label{lemma_random}
\end{lemma}

When the data does not align along any particular directions, $n_0$ will be much smaller than $n_2$ since the vector $\bc_2$ can only be simultaneously orthogonal to a small number of the columns of $\bD_2$. 
Noting that the LHS of (\ref{random_suf}) has order $n_1$ and the RHS has order $\sqrt{n_2}+n_0$ (which is much smaller than $n_2$ when $n_2$ is sufficiently large), we see that the sufficient conditions are naturally satisfied when the data is well-distributed within the subspaces.

\subsection{Clustering multiple subspaces}
\label{sec:multiple}
In this section, the performance guarantees provided in Theorem \ref{two_ind} are extended to  more than two subspaces. 
iPursuit identifies the subspaces consecutively, i.e., one subspace is identified in each step. The data lying in the identified subspace is removed and optimal direction-search (\ref{convex}) is applied to the remaining data points to find the next subspace. This process is continued to find all the subspaces.
%
%
In order to analyze iPursuit for the scenarios with more than two subspaces, it is helpful to define the concept of minimum innovation. 
\begin{definition}
$(\bD_i , \calS_i)$ is said to have minimum innovation w.r.t. the vector $\bq$ in the set $\left\{ (\bD_j , \calS_j)  \right\}_{j=1}^m$ if and only if
\begin{eqnarray}
\underset{\bc \in \mathcal{I} \big( \mathcal{S}_i \perp {\overset{m}{\underset{k = 1 \atop k \neq i}{\oplus}}} \mathcal{S}_k \big) \atop \bc^T \bq = 1}{\inf} \: \:  \| \bc^T \bD_i \|_1 \: < \underset{\bc \in \mathcal{I} \big( \mathcal{S}_j \perp {\overset{m}{\underset{k = 1 \atop k \neq j}{\oplus}}} \mathcal{S}_k \big) \atop \bc^T \bq = 1}{\inf} \: \:  \| \bc^T \bD_j \|_1
\label{stepclear}
\end{eqnarray}
for every $ j \neq i \: , \: 1 \leq j \leq m \:$, where $\bq$ is a unit $\ell_2$-norm in $\oplus_{k = 1}^m \calS_k$.
\end{definition}

If $(\bD_k , \calS_k)$ has minimum innovation in the set $\left\{ (\bD_j , \calS_j)  \right\}_{j=1}^N$ (w.r.t. the vector $\bq$ used in the first step), then we expect iPursuit to find $\calS_k$ in the first step. Similar to Theorem \ref{two_ind}, we make the following assumption without loss of generality.

\begin{assumption}
Assume that $(\bD_k , \calS_k)$ has minimum innovation w.r.t. $\bq_k$ in the set $\left\{ (\bD_j , \calS_j)  \right\}_{j=1}^k \:$  for $2 \leq k \leq N$.
\label{asm_order}
\end{assumption}

According to Assumption \ref{asm_order}, if $\bq_N$ is used in the first step as the linear constraint of the innovation pursuit optimization problem, iPursuit is expected to first identify $\calS_N$. 
In each step, the problem is equivalent to disjoining two subspaces. In particular, in the ${(N-m+1)}^{\text{th}}$ step, the algorithm is expected to identify $\calS_m$, which can be viewed as separating $(\bD_m , \calS_m)$ and $(\overset{m-1} {\underset{i = 1 }{\cup}} \bD_i \: , \overset{m-1} {\underset{i = 1 }{\oplus}} \mathcal{S}_i )$ by solving  $(\text{IP}_m)$
\begin{eqnarray}
(\text{IP}_m) \hspace{.5cm}\begin{aligned}
& \underset{\hat{\bc}}{\min}
& & \bigg\| \hat{\bc}^T \left( \overset{m}{\underset{k = 1 }{\cup}} \bD_k \right) \bigg\|_1\\
& \text{subject to}
& & \hat\bc \in  \overset{m}{\underset{k = 1 }{\oplus}} \calS_k \: \: , \: \:   \hat{\bc}^T \bq_m = 1 \:.
\end{aligned}
\label{eq:OP}
\end{eqnarray}
Note that $\overset{m}{\underset{k = 1 }{\oplus}} \calS_k$ is the span of the data points that have not been yet removed.
Based on this observation, we can readily state the following theorem, which provides sufficient conditions for iPursuit to successfully identify $\calS_m$ in the ${(N-m+1)}^{\text{th}}$ step. The proof of this theorem follows directly from Theorem \ref{two_ind} with two subspaces, namely, $\calS_m$ and $\overset{m-1} {\underset{i = 1 }{\oplus}} \mathcal{S}_i$. Similar to Theorem \ref{two_ind}, the sufficient conditions are characterized in terms of the optimal solution to an oracle optimization problem ($\text{OP}_m)$ defined below.
\begin{eqnarray}
(\text{OP}_m) \hspace{0.5cm} \begin{aligned}
& \underset{\hat{\bc}}{\min}
& & \| \hat{\bc}^T \bD_m \|_1\\
& \text{subject to}
& & \bc \in \mathcal{I} \big( \mathcal{S}_m \perp \overset{m-1}{\underset{k = 1 }{\oplus}} \calS_k \big) ,~ \hat{\bc}^T \bq_m = 1 \:.
\end{aligned}
\label{eq:Pm}
\end{eqnarray}
We also define
$
\calL_{0m} = \{i\in[n_m]: ~\bc_m^T \bd_i = 0, ~\bd_i \in \bD_m\}
$ with cardinality $n_{0m}$,
where $\bc_m$ is the optimal point of (\ref{eq:Pm}).

\begin{theorem}
Suppose that the data follows Data model 1 with $N = m$ and assume that $\bD_m$ cannot follow Data Model 1 with $N > 1$.  Assume that $(\bD_m , \calS_m)$ has minimum innovation with respect to $\bq_m$ in the set $\left\{ (\bD_j , \calS_j)  \right\}_{j=1}^m \:$. Define $\bc_m$ as the optimal point of $(\text{OP}_m)$ in (\ref{eq:Pm}) and let
$
\alpha_m = \sum_{\bd_i \in \bD_m }^{} \sgn (\bc_m^T \bd_i) \bd_i
$. 
Assume  $\bq_m$ is a unit $\ell_2$-norm vector in $\overset{m}{\underset{k = 1 }{\oplus}} \calS_k \: $.
If
\begin{eqnarray}
\begin{aligned}
& \frac{1}{2} \underset{\delta \in \overset{m-1}{\underset{k = 1 }{\oplus}} \calS_k  \atop \| \delta \| = 1}{\inf} \sum_{\bd_i \in \overset{m-1}{\underset{k = 1 }{\cup}} \bD_k} \hspace{-.3cm}\left| \mathbf{\delta}^T \bd_i \right| >  \| \bV_m^T \bT_{m-1} \| \bigg( \|\alpha_m \| + n_{0m} \bigg), \\
& \frac{ \| \bq_m^T \bP_m \|}{ 2\| \bq_m^T \bT_{m-1} \|}  \left( \underset{\delta \in \overset{m-1}{\underset{k = 1 }{\oplus}} \calS_k  \atop \| \delta \| = 1}{\inf} \sum_{\bd_i \in \overset{m-1}{\underset{k = 1 }{\cup}} \bD_k} \left| \mathbf{\delta}^T \bd_i \right|  \right) \\
&\qquad\qquad\qquad\qquad >   \| \bV_m^T \bP_m \| \bigg( \| \alpha_m \| + n_{0m} \bigg),
\end{aligned}
\label{totalsuffc}
\end{eqnarray}
where $\bT_{m-1}$ is an orthonormal basis for  $\overset{m-1}{\underset{k = 1 }{\oplus}} \calS_k$ and $\bP_m$ is an orthonormal basis for $\mathcal{I} \big( \mathcal{S}_m \perp \overset{m-1}{\underset{k = 1 }{\oplus}} \calS_k \big)$, then $\bc_m$, which lies in $\mathcal{I} \big( \mathcal{S}_m \perp \overset{m-1}{\underset{k = 1 }{\oplus}} \calS_k \big)$, is the optimal point of $(\text{IP}_m)$ in (\ref{eq:OP}),
i.e., the subspace $\calS_m$ is correctly identified.
 \label{corol1}
\end{theorem}

The sufficient conditions provided in Theorem \ref{corol1} reveal another intriguing property of iPursuit. Contrary to conventional wisdom, increasing the number of subspaces may improve the performance of iPursuit, for if the data points are well distributed in the subspaces, the LHS of (\ref{totalsuffc}) is more likely to 
dominate the RHS. 
In the appendix section, we further investigate the  sufficient conditions (\ref{totalsuffc}) and simplify the LHS to the permeance statistic.


\subsection{Complexity analysis}
\label{sec:comp_analysis}
In this section, we use an Alternating Direction
Method of Multipliers (ADMM) \cite{boyd2011distributed} to develop an
efficient algorithm for solving (IP).
Define $\bU \in \mathbb{R}^{M_1 \times r}$ as an orthonormal basis for $\calD$, where $r$ is the rank of $\bD$. Thus, the optimization problem (\ref{convex}) is equivalent to
$
\underset{\ba}{\min} \:\:
 \| \ba^T \bU^T \bD \|_1 \quad \text{subject to} \quad\ba^T \bU^T \bq = 1
$. Define 
$\mathbf{f} = \bU^T \bq$ and $\bF = \bU^T \bD$. Hence, this optimization problem is equivalent to
\begin{eqnarray}
\begin{aligned}
& \underset{\ba , \bt}{\min}
& & \| \bt \|_1 + \frac{\mu}{2} \|\bt - \bF^T \ba\|^2 + \frac{\mu}{2} ( \ba^T \mathbf{f} - 1 )^2  \\
& \text{subject to}
& & \bt = \bF^T \ba \: \: , \: \: \ba^T \mathbf{f} = 1 \:,
\end{aligned}
\label{eq:adm1}
\end{eqnarray}
with a regularization parameter $\mu$.
The Lagrangian function of
(\ref{eq:adm1}) can be written as
\begin{eqnarray}
\begin{aligned}
\calL(\bt , \ba , \by_1 , y_2) =
 & \| \bt \|_1 + \frac{\mu}{2} \| \bF^T \ba - \bt \|^2 + \frac{\mu}{2} ( \ba^T \mathbf{f} - 1 )^2  \\
&  \by_1^T  ( \bF^T \ba - \bt) +  y_2(\ba^T \mathbf{f} - 1) \:,
\end{aligned}
\label{eq:adm2}
\end{eqnarray}
where $\by_1$ and $y_2$ are the Lagrange multipliers. The ADMM approach uses an iterative procedure. Define $(\ba_k , \bt_k)$ as the optimization variables and $(\by_1^{k} , y_2^{k})$  the Lagrange multipliers at  the $k^{\text{th}}$ iteration. Define $\bG := \mu^{-1}(\bF \bF^T + \mathbf{f}\mathbf{f}^T)^{-1}$ and the element-wise
function 
$\calT_{\epsilon}(x) := \sgn(x) \max ( |x| - \epsilon , 0 )$. Each iteration consists of the following
steps:

\noindent
1. Obtain $\ba_k$ by minimizing the Lagrangian function with
respect to $\ba$ while the other variables are held constant. The
optimal $\ba$ is obtained as
\begin{eqnarray}
\ba_{k+1} = \bG \big(\mu \bF \bt_{k} - \bF \by_1^{k} + \mathbf{f} (\mu - y_2^{k}) \big) \:. \nonumber
\end{eqnarray}

\smallbreak
\noindent
2. Similarly, update $\bt$ as
\begin{eqnarray}
\bt_{k+1} = \calT_{\mu^{-1}} ( \bF^T \ba_{k+1} + \mu^{-1} \by_1^{k} ) \:.
\nonumber
\end{eqnarray}

\smallbreak
\noindent
3. Update the Lagrange multipliers as follows
\begin{eqnarray}
\begin{aligned}
& \by_1 = \by_1 + \mu (\bF^T \ba_{k+1} - \bt_{k+1}) \: \: , \: \: y_2^{k+1} = y_2^{k} + \mu (\ba^T \mathbf{f} - 1) \:.
\nonumber
\end{aligned}
\end{eqnarray}

\noindent
These 3 steps are repeated until the algorithm converges or
the number of iterations exceeds a predefined threshold.
The complexity of the initialization step of the solver is $\calO(r^3)$ plus the complexity of obtaining $\bU$.
Obtaining an appropriate $\bU$ has $O(r^2 M_2)$ complexity by applying the clustering algorithm to a random subset of the rows of $\bD$ (with the rank of sampled rows equal to $r$).
In addition, the complexity of each iteration of the solver is $\calO (r M_2)$. Thus, the overall complexity is less than $\calO ((r^3 + r^2 M_2)N)$ since the number of data points remaining keeps decreasing over the iterations. In most cases, $r \ll M_2$, hence the overall complexity is roughly $\calO(r^2 M_2 N)$.

iPursuit brings about substantial speedups over most existing algorithms due to the following: i) Unlike existing \textit{iterative} algorithms (such as RANSAC) which have exponential complexity in the number and dimension of subspaces, the complexity of iPursuit is linear in the number of subspaces and quadratic in their dimension. In addition, while iPursuit has linear complexity in $M_2$, spectral clustering based algorithms have complexity $\calO(M_2^2 N)$ for their spectral clustering step plus the complexity of obtaining the similarity matrix,
 ii) More importantly, the solver of the proposed optimization problem has $\calO(r M_2)$ complexity per iteration, while the other operations -- whose complexity are $\calO(r^2 M_2)$ and $\calO(r^3)$ -- sit outside of the iterative solver. This feature makes the proposed method notably faster than most existing algorithms which solve high-dimensional optimization problems.  For instance, solving the optimization problem of the SSC algorithm has roughly $\calO (M_2^3 + r M_2)$ complexity per iteration \cite{elhamifar2013sparse}.

\subsection{How to choose the vector $\bq$?}
\label{sec:choose_q}
The previous analysis revealed that the coherency of the vector $\bq$ with the innovation subspace is a key performance factor for iPursuit.
While our investigations have shown that the proposed algorithm performs very well when the subspaces are independent even when the vector $\bq$ is chosen at random,
randomly choosing the vector $\bq$ may not be favorable when the dimension of their intersection is increased (c.f. Section \ref{sec:2sub_perf}).
This motivates the methodology described next that aims to identify a ``good'' choice of the vector $\bq$.

Consider the following least-squares optimization problem,
\begin{align}
 \underset{\hat{\bq}}{\min}
 \| \hat{\bq}^T \bD \|_2
 \quad \text{s. t.} \quad
\hat{\bq} \in \mathcal{D} \quad \text{and} \quad  \| \hat{\bq} \| = 1.
\label{norm_2}
\end{align}
The optimization problem (\ref{norm_2}) searches for a vector in $\calD$ that has a small projection on the columns of $\bD$.
The optimal point of (\ref{norm_2}) has a closed-form solution, namely, the singular vector corresponding to the least non-zero singular value of $\bD$.  When the subspaces are close to each other, the optimal point of (\ref{norm_2}) is very close to the innovation subspace $\calI \left( \calS_2 \perp \calS_1 \right)$. This is due to the fact that $\calI \left( \calS_2 \perp \calS_1 \right)$ is orthogonal to $\calS_1$, hence a vector in the innovation subspace will have a small projection on $\calS_2$.
%
As such, when the subspaces are close to each other, the least singular vector is coherent with the innovation subspace and can be a good candidate for the vector $\bq$. In the numerical results section, it is shown that this choice of $\bq$ leads to substantial improvement in performance compared to using a randomly generated $\bq$. However, in settings in which the singular values of $\bD$ decay rapidly and the data is noisy we may not be able to obtain an exact estimate of $r$. This may lead to the undesirable usage of a singular vector corresponding to noise as the constraint vector.  In the next section, we investigate stability issues and present robust variants of the algorithm in the presence of noise. We remark that with real data or when the data is noisy, by the least singular vector we refer to \emph{the least dominant} singular vector and not to the one corresponding to the smallest singular value which is surely associated with the noise component.

\section{Noisy data}
\label{sec:noisy}
\begin{figure}[t!]
 \centering
    \includegraphics[width=0.50\textwidth]{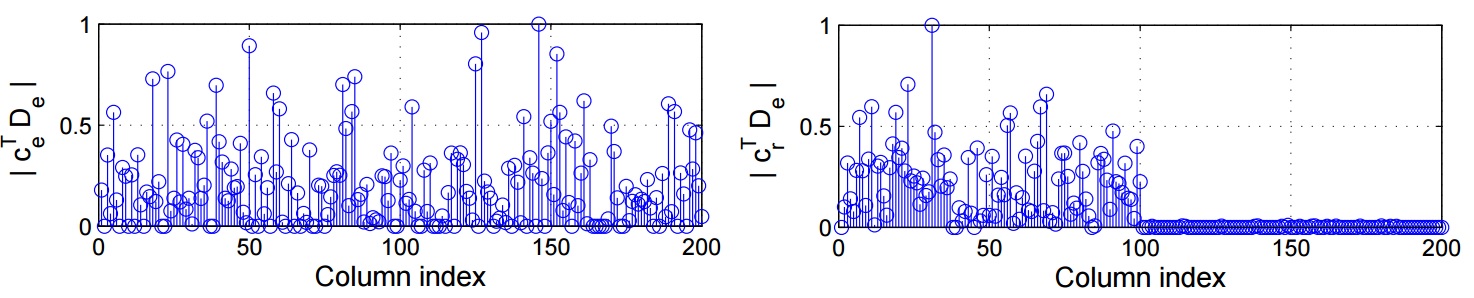}
    \vspace{-0.7cm}
    \caption{The left plot shows the output of (\ref{noisyone}), while the right plot shows the output of iPursuit when its search
domain is restricted to the subspace spanned by the dominant singular
vectors as per (\ref{eq:robust_noise}).  }
    \label{fig:with noise}
\end{figure}

In the presence of additive noise, we model the data as
\begin{eqnarray}
\bD_e = \bD + \bE \: ,
\label{eq:with noise}
\end{eqnarray}
where $\bD_e$ is the noisy data matrix, $\bD$ the clean data which follows Data model 1 and $\bE$ the noise component. 
The rank of $\bD$ is equal to $r$. Thus, the singular values of $\bD_e$ can be divided into two subsets: the dominant singular values (the first $r$ singular values) and the small singular values (or the singular values corresponding to the noise component).  Consider the optimization problem (IP) using $\bD_e$, i.e.,
\begin{align}
 \underset{\hat{\bc}}{\min}\:\:
 \| \hat{\bc}^T \bD_e \|_1
 \quad \text{s.t.} \quad
  \hat{\bc} \in \spn(\bD_e) \quad \text{and} \quad   \hat{\bc}^T \bq = 1.
  \label{noisyone}
\end{align}
Clearly, the optimal point of (\ref{noisyone}) is very close to the subspace spanned by the singular vectors corresponding to the small singular values. Thus, if $\bc_e$ denotes the optimal solution of (\ref{noisyone}), then all the elements of $\bc_e^T \bD_e$ will be fairly small and we cannot distinguish the subspaces.
However, the span of the dominant singular vectors is approximately equal to $\calD$. Accordingly, we propose the following approximation to (IP),
\begin{align}
  \underset{\hat{\bc}}{\min}  \:\:
  \| \hat{\bc}^T \bD_e \|_1
  \quad \text{s. t.} \quad
  \hat{\bc} \in \spn(\bQ) \quad \text{and} \quad   \hat{\bc}^T \bq = 1,
  \label{eq:robust_noise}
\end{align}
%
where $\bQ$ is an orthonormal basis for the span of the dominant singular vectors. The first constraint of (\ref{eq:robust_noise}) forces the optimal point to lie in $\spn(\bQ)$, which serves as a good approximation to $\spn(\bD)$. For instance, consider $\bD = [\bD_1 ~ \bD_2]$, where the columns of $\bD_1 \in \mathbb{R}^{40 \times 100}$ lie in a 5-dimensional subspace $\calS_1$, and the columns of $\bD_2 \in \mathbb{R}^{40 \times 100}$ lie in another 5-dimensional subspace $\calS_2$. Define $\bc_e$ and $\bc_r$ as the optimal points of (\ref{noisyone}) and (\ref{eq:robust_noise}), respectively. Fig. \ref{fig:with noise} shows $| \bc_e^T \bD_e |$ and $| \bc_r^T \bD_e |$ with the maximum element scaled to one.
Clearly, $\bc_r^T \bD_e$ can be used to correctly cluster the data. \textcolor{black}{ In addition, when $\bD$ is low rank, the subspace constraint in (\ref{eq:robust_noise}) can filter out a remarkable portion of the noise component.}

When the data is noisy and the singular values of $\bD$ decay rapidly, it may be hard to accurately estimate $r$. If the dimension is incorrectly estimated, $\bQ$ may contain some singular vectors corresponding to the noise component, wherefore the optimal point of (\ref{eq:robust_noise}) could end up near a noise singular vector. In the sequel, we present two  techniques to effectively avoid this undesirable scenario.  

\smallbreak
\noindent 1. \textit{Using a data point as a constraint vector:}
A singular vector corresponding to the noise component is nearly orthogonal to the entire data, i.e., has small projection on all the data points. Thus, if the optimal vector is forced to have strong projection on a data point, it is unlikely to be close to a noise singular vector. Thus, we modify (\ref{eq:robust_noise}) as
\begin{align}
 \underset{\ba }{\min}
 \| \ba^T \bQ^T \bD_e \|_1 \quad \text{s.t.}  \quad   \ba^T \bQ^T \bq = 1,~\text{and}~ \bq = {\bd_e}_k \:,
 \label{eq:robust2}
\end{align}
where 
${\bd_e}_k$ is the $k^{\text{th}}$ column of $\bD_e$. The modified constraint in (\ref{eq:robust2}) ensures that the optimal point is not orthogonal to ${\bd_e}_k$. If ${\bd_e}_k$ lies in the subspace $\calS_i$, the optimal point of (\ref{eq:robust2}) will lie in the innovation subspace corresponding to $\calS_i$ whp. To determine a good data point for the constraint vector, we leverage the principle presented in section \ref{sec:choose_q}. Specifically, we use the data point closest to the least dominant singular vector rather than the least dominant singular vector itself.

\smallbreak
\noindent 2. \textit{Sparse representation of the optimal point:}
When $\bD$ is low rank, i.e., $r \ll \min(M_1,M_2)$, any direction in the span of the data -- including the optimal direction sought by iPursuit -- can be represented as a sparse combination of the data points.
For such settings, we can rewrite (\ref{eq:robust2}) as
\begin{eqnarray}
\begin{aligned}
& \underset{\ba , \bz}{\min}
& & \| \ba^T \bQ^T \bD_e \|_1 + \gamma \|\bz\|_1  \\
& \text{subject to}
& & \ba = \bQ^T \bD_e \: \bz \quad \text{and} \quad \ba^T \bQ^T {\bd_e}_k = 1 \:,
\end{aligned}
\label{eq: sparse rep}
\end{eqnarray}
where $\gamma$ is a regularization parameter.
 Forcing a sparse representation in (\ref{eq: sparse rep}) for the optimal direction averts a solution that lies in close proximity with the small singular vectors, which are normally obtained through linear combinations of a large number of data points. Using the data points as constraint vectors effectively accords robustness to the imperfection in forming the basis matrix $\bQ$. However, our investigations show that enforcing the sparse representation for the optimal direction can enhance the performance in some cases. 
\textcolor{black}{The table of Algorithm 1 details the proposed method for noisy data along with used notation and definitions.} 


\begin{algorithm}
\caption{Innovation pursuit for noisy data (the iterative framework)}
{\footnotesize
\textbf{Initialization} Set $\hat{n}$ and $\hat{N}$ as integers greater than 1, and set $c_i$ and $c_o$ equal to positive real numbers less than 1.
\smallbreak
\textbf{While} The number of identified subspaces is less than $\hat{N}$ or the number of columns of $\bD_e$ is greater than $\hat{n}$.
\smallbreak
 \textbf{1. Obtaining the basis for the remaining Data:}  Construct $\bQ$ as the orthonormal matrix formed by the dominant singular vectors of $\bD_e$.
\smallbreak
\textbf{2. Choosing the vector $\bq$:} Set $\bq = $ the column of $\bD_e$ closest to the last column of $\bQ$.
\smallbreak
\textbf{3. Solve}  (\ref{eq: sparse rep}) and
define $\bc^{*} = \bQ \ba^*$, where $\ba^*$ is the optimal point  and define $\bh_1 = \frac{ \big |\bD_e^T \bc^{*} \big |}{\max (\big| \bD_e^T \bc^{*} \big|)}$.
\smallbreak
\textbf{4. Finding a basis for the identified subspace: }
Construct the matrix $\bG_1$ from the columns of $\bD_e$ corresponding to the elements of $\bh_1$ greater than $c_i$. Define matrix $\bF_1$ as a orthonormal basis for the dominant left singular vectors of $\bG_1$. 

\smallbreak
\textbf{5. Finding a basis for the rest of the data:}
Define the vector $\bh_2$ whose entries are equal to the $\ell_2$-norm of the columns of $(\bI - \bF_1 \bF_1^T) \bD_e$.  Normalize $\bh_2$ as $\bh_2 := \bh_2/ \max(\bh_2)$. Construct $\bG_2$ as the columns of $\bD_e$ corresponding to the elements of $\bh_2$ greater than $c_o$. Define $\bF_2$ as an orthonormal basis for the columns of $\bG_2$.
\smallbreak
\textbf{6. Find the data point belonging to the identified subspace:} Assign ${\bd_e}_i$ to the identified subspace if $\| \bF_1^T {\bd_e}_i \| \ge \| \bF_2^T {\bd_e}_i \|$.
\smallbreak
\textbf{7. Remove the data points belonging to the identified subspace:} Update $\bD_e$ by removing the columns corresponding to the identified subspace.
\smallbreak
\textbf{End While}
}
\end{algorithm}

\begin{remark}
\textcolor{black}{
The proposed method can be made parameter-free if we can avoid thresholding in Steps 4 and 5 of Algorithm 1. Indeed, in Step 4 we can construct $\bG_1$ using the columns of $\bD_e$
corresponding to the $\kappa$ largest elements of $\bh_1$. $\kappa$ has to be chosen large enough such that the sampled columns span the identified subspace, and hence can be set if we have access to an upper bound on the dimension of the subspaces, which is naturally available in many applications.
For example, in motion segmentation we know that $r_i \leq 4$. The matrix $\bG_2$ can be constructed in a similar way, i.e., without thresholding.}
\end{remark}

\subsection{Minimizing error propagation}
If $\kappa$ (or $c_i$) and the threshold $c_o$ in Algorithm 1 are chosen appropriately, the algorithm exhibits strong robustness in the presence of noise. Nonetheless, if the data is too noisy, an error incurred in one step of the algorithm may propagate and unfavorably affect the performance in subsequent steps. In the following, we discuss the two main sources of error and present some techniques to effectively neutralize their impact on subsequent iterations.

\smallbreak
\noindent \textbf{A.1} \textit{Some data points are erroneously included in $\bG_1$ and $\bG_2$:} Suppose that $\calS_m$ is the subspace to be identified in a given step of the algorithm, 
 i.e., the optimal point of (\ref{eq: sparse rep}) lies in the innovation subspace corresponding to $\calS_m$. If the noise component is too strong, few data points from the other subspaces may be erroneously included in $\bG_1$. In this subsection, we present a technique to remove these erroneous data points (In \cite{rahmnia2016coh}, we analyze the proposed technique as a robust subspace recovery algorithm).
Consider two columns $\bg_1$ and $\bg_2$ of $\bG_1$, where $\bg_1$ belongs to $\calS_m$ and $\bg_2$ to one of the other subspaces, and define the inner products $\alpha_1 := \| \bg_1^T \bG_1 \|$ and $\alpha_2 := \| \bg_2^T \bG_1 \|$.
%
Since $\bG_1$ contains many data points that are coherent with $\bg_1$, $\alpha_1 > \alpha_2$ whp. Thus, by removing a portion of the columns of $\bG_1$ with small inner products, the columns belonging to the other subspaces are likely to be removed.
In addition, we obtain $\bF_1$ from the principal directions of $\bG_1$ which mitigates the impact of noise and erroneous data points. The same technique can be used to remove the wrong data points from $\bG_2$. The table of
\textcolor{black}{
Algorithm 2 presents the details of using the proposed idea in the fourth step of Algorithm 1 to remove the erroneous data points from $\bG_1$. The complexity of this extra step is roughly $\calO(M_2^2 r)$. The proposed method is remarkably faster than the state-of-the-art subspace clustering algorithms even with this additional step since the complexity of solving the underlying optimization problem is linear in the number of data points. In section \ref{sec:running}, we compare the run time of the proposed method to that of the state-of-the-art algorithms.
}

\begin{algorithm}
\caption{Fourth step of Algorithm 1 with a technique for removing erroneous data points}
{\footnotesize
\textbf{Initialization} Set $\beta$ equal to an integer between 0 and 50. \\
\textbf{4. Finding a basis for the identified subspace}\\
\textbf{4.1} Construct the matrix $\bG_1$ from the columns of $\bD_e$ corresponding to the elements of $\bh_1$ that are greater than $c_i$, or using the columns of $\bD_e$ corresponding to the $\kappa$ largest elements of $\bh_1$. \\
\textbf{4.2} Define $\bR = \bG_1^T \bG_1$. Remove $\beta$ percent of the columns of $\bG_1$ corresponding to the columns of $\bR$ with the smallest $\ell_2$-norms. \\
\textbf{4.3} Define $\bF_1$ as an orthonormal basis for $\bG_1$.
}
\end{algorithm}

\begin{figure}[t!]
 \centering
    \includegraphics[width=0.44\textwidth]{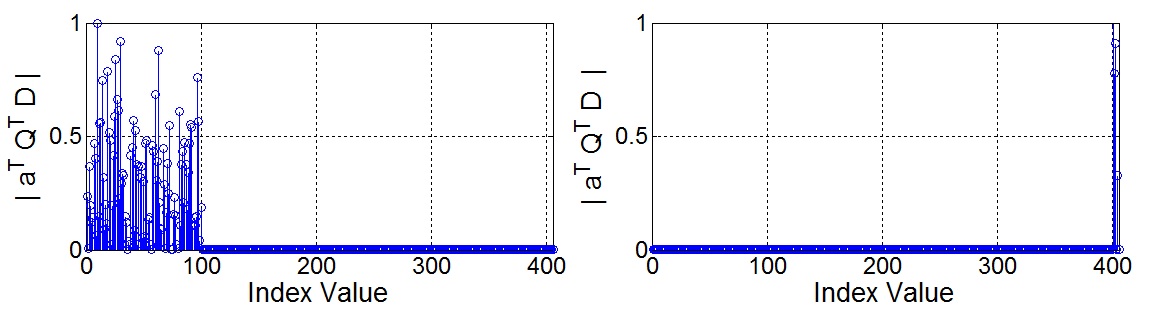}
    \vspace{-0.3cm}
    \caption{  \textcolor{black}{ The output of the proposed method with different choices of the constraint vector. In the left and right plots, the first and the $401^{\text{th}}$ column are used as a constraint vector, respectively. The first column lies in a cluster with 100 data points and the $401^{\text{th}}$ column lies in a cluster with 6 data points. }  }
    \label{fig: error_prop}
\end{figure}

\smallbreak
\noindent \textbf{A.2} \textit{Some of the data points remain unidentified:} Suppose $\calS_m$ is to be identified in a given iteration, yet not all the data points belonging to $\calS_m$ are identified, i.e., some of these points remain unidentified.
In this case, an error may occur if one such point is used for the constraint vector $\bq$.
However, such an error can be easily detected because if one such point is used as $\bq$, the vector $\bh_1$ would be too sparse since the optimal direction is orthogonal to all the data points expect a few remaining points of $\calS_m$.
As an example, consider the setting where $\bD$ follows Data model 1 with $N=5$, $\{r_i\}_{i=1}^5 = 5$, $\{n_i \}_{i=1}^4 = 100$ but $n_5 = 6$, i.e., $\calS_5$ contains only few data points. Fig. \ref{fig: error_prop} shows the output of (\ref{eq:robust2}) with $\bq = \bd_1$ and $\bq = \bd_{401}$. The right plot shows a solution that is too sparse. Accordingly, if the output of (\ref{eq:robust2}) is too sparse, we solve (\ref{eq:robust2}) again using a new constraint vector.

\smallbreak
\noindent \textbf{A.3} \textit{Error correction:}
The robustness of spectral clustering based algorithms stems from the fact that the spectral clustering step considers the representation of each of the data points, thus few errors in the similarity matrix do not significantly impact the overall performance of the algorithm.
The proposed multi-step algorithm is of low complexity, however, if a data point is assigned to an incorrect cluster in a given step, its assignment will not change in subsequent steps of the algorithm that only consider the remaining data points.
Thus, for scenarios where the data is highly noisy, we propose a technique for final error correction to account for and correct wrong assignments of data points to incorrect clusters. The table of Algorithm 3 presents the proposed technique, which is applied to the clustered data after Algorithm 1 terminates 
to minimize the clustering error. It uses the idea presented in Algorithm 2 to obtain a set of
bases for the subspaces with respect to which the clustering is subsequently updated. In fact, Algorithm 3 can be applied multiple times, each time updating the clustering.

\subsection{Using multiple constraint vectors}
In each step of the proposed algorithm, we could solve (\ref{eq: sparse rep}) with multiple choices for $\bq$ and pick the one leading to the most favorable solution. For instance, we can find the $m$ nearest neighbors to the least dominant singular vector among the data points, or find $m$ data points that are most close to the $m$ least dominant singular vectors, and solve (\ref{eq:robust2}) with all the $m$ choices of the constraint vector. The question remains as of how to identify the best choice for $\bq$. Ideally, one would favor the constraint vector that minimizes the clustering error.
Note that each step of the proposed algorithm is clustering the data points into two subspaces. \emph{When the clustering error increases, the distance between the identified subspaces decreases}. To clarify, consider $\bD_1$ and $\bD_2$ spanning subspaces $\calS_1$ and $\calS_2$, respectively. A subspace clustering algorithm clusters the data matrix $\bD = [\bD_1 \: \: \bD_2]$ into $\bD_1^{'}$ and $\bD_2^{'}$.
If the number of data points belonging to $\calS_2$ in $\bD_1^{'}$ increases, the identified subspace corresponding to $\calS_1$ gets closer to $\calS_2$. 
As such, we choose the constraint vector which leads to the maximum distance between the identified subspaces. The distance between two subspaces can be measured using their orthogonal bases. For instance, $\| \bV_1^T \bV_2 \|_F^2$ can be used as a distance measure between $\calS_1$ and $\calS_2$ \cite{park2014greedy,soltanolkotabi2012geometric}, as it is inversely proportional to the distance between $\calS_1$ and $\calS_2$. 

\begin{algorithm}
\caption{Final error correction}
{\footnotesize
\
\noindent\textbf{Define} the matrices $\{ \hat{\bD}_i \}_{i = 1}^{\hat{N}}$ as the clustered data.

\smallbreak
\textbf{Error Correction}\\
\textbf{1} \textbf{For} $ 1 \leq i \leq \hat{N}$

\textbf{1.1} Define $\bR_i = \hat{\bD}_i^T \hat{\bD}_i$. Remove $\beta$ percent of the columns of $\hat{\bD}_i$ corresponding to the columns of $\bR_i$ with the smallest $\ell_2$-norms.

\textbf{1.2} Obtain $\hat{\bV}_i$ as an orthonomal basis for the column space of $\hat{\bD}_i$.\\
\textbf{End For}

\smallbreak
\textbf{2} Update the data clustering with respect to the obtained bases $\{ \hat{\bV}_i \}_{i = 1}^{\hat{N}}$ (the matrices $\{ \hat{\bD}_i \}_{i = 1}^{\hat{N}}$ are updated), i.e., a data point $\bd$ is assigned to the $i^{\text{th}}$ cluster if $ i = \underset{k}{ \argmax} \: \| \bd^T \hat{\bV}_k \|_2$.
\\

}
\end{algorithm}


\section{ \textcolor{black}{Innovation pursuit with spectral clustering}}
\label{sec:integr}
In Section \ref{sec:proposed_secd}, we have shown analytically that the direction of innovation is a powerful tool for distinguishing subspaces. The presented iterative algorithm, Algorithm 1, uses the directions of innovation iteratively to consecutively identify the subspaces. Algorithm 1 is provable, fast, and strongly robust to the intersection between the subspaces but has some limitations. 
In this section, we first discuss the limitations of the proposed iterative approach. Then, we show that the proposed innovation search method can be integrated with spectral clustering to yield a new spectral-clustering-based subspace segmentation algorithm that does not have the limitations of Algorithm 1.

\subsection{Limitations of Algorithm 1}
\label{sec:restrcition}

\begin{figure}[t!]
	\centering
    \includegraphics[width=0.47\textwidth]{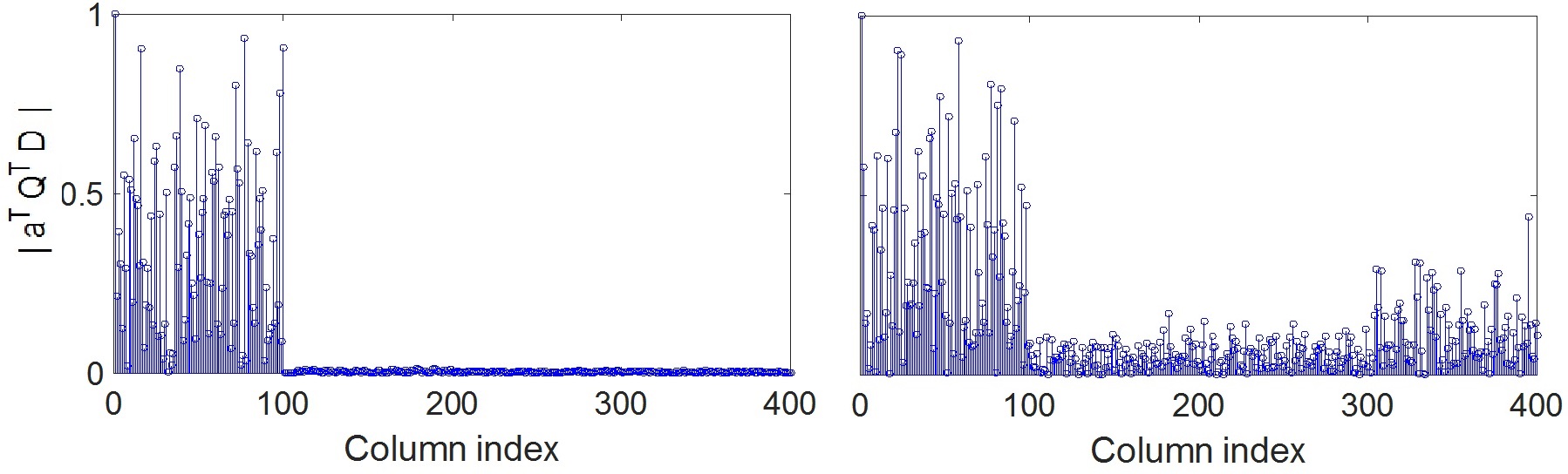}
    \vspace{-.1in}
    \caption{ \textcolor{black}{ The data lies in a union of 4 random 5-dimensional subspaces. 
In the left plot $M_1 = 30$. In the right plot, $M_1 = 10$. For both plots, the first column is used as a constraint vector.} }
    \label{fig: inov req}
\end{figure}

\subsubsection{Innovation}
The main requirement of Algorithm 1 is that no subspace should lie in the direct sum of the other subspaces.
In other words, the dimension of the innovation subspace corresponding to each subspace should be at least 1. Thus, the number of subspaces cannot be larger than the ambient dimension.
More specifically, suppose the data lies in a union of $N$ $d$-dimensional subspaces and that the dimension of the innovation subspace corresponding to each subspace is equal to $g$. In this case, the dimension of the ambient space should be greater than or equal to $d + g(N-1)$.
As an example, assume the data points lie in a union of 4 random 5-dimensional subspaces, i.e., $\bD = [\bD_1 \: \: \bD_2 \: \: \bD_3 \: \: \bD_4]$, each with 100 data points. Fig. \ref{fig: inov req} shows $\ba^T \bQ^T \bD$ with $\bq = \bd_1$. In the left and right plots $M_1 = 30$ and $M_1  = 10$, respectively. When $M_1 = 30$, the dimension of the innovation subspace corresponding to each subspace is equal to 5 whp. But, when $M_1 = 10$, not every subspace carries innovation relative to the other subspaces. 
 In the left plot, the non-zero elements correspond to only one subspace, and also the span of the data points corresponding to the zero elements does not contain the span of those corresponding to the non-zero elements. Thus, Algorithm 1 can successfully identify and isolate the first subspace. By contrast, both of these conditions are violated in the scenario of the right plot. 

\subsubsection{Sub-clusters with innovation}
According to the second condition of Lemma 1, the data points in a cluster should not  
lie in the union of lower dimensional subspaces each with innovation w.r.t. to the other ones. For instance, assume $\bD_1 = [\bD_1 \:\:\bD_2]$ and $\bD_1 = [\bD_{11} \:\: \bD_{12}]$. The data points in $\bD_{11}$, $\bD_{12}$, and $\bD_2$ span three independent 1-dimensional subspaces. In this case, Algorithm 1 will identify three subspaces since the column spaces of $\bD_{11}$ and $\bD_{12}$ have innovation relative to the other subspaces. 
If it is sensible to cluster the data into three clusters, then the algorithm will yield correct clustering in that sense. However, if it only makes sense to cluster the data into exactly two clusters, then the algorithm may not cluster the data correctly. For example, one possible output would be $\{\bD_{11} \: , \: [\bD_{12} \: \: \bD_2] \}$.

\subsection{Integration with spectral clustering}
The optimization problem (\ref{eq:robust2}) finds a direction in the span of the data that has large projection on the $k^{\text{th}}$ data point and small projections on the other data points. In practice, the data points within a subspace are mutually coherent, wherefore the optimal point of (\ref{eq:robust2}) will naturally have strong projection on other data points in the subspace containing ${\bd_e}_k$. The fact that the obtained direction of innovation is simultaneously strongly coherent with the $k^{\text{th}}$ data point -- and in turn some of its neighbors -- and highly incoherent with the other subspaces makes it a particularly powerful tool to identify a set of data points belonging to the same subspace containing the $k^{\text{th}}$ point as we show in detail in \cite{rahmani2017direction}.
For instance, even with the unwieldy scenario of the right plot of Fig. \ref{fig: inov req}, the few data points that correspond to the elements with the largest values all lie in the same subspace. 
As such, by solving the innovation search optimization problem (\ref{eq:robust2}) for each of the data points, $\{ {\bd_e}_k \}_{i=1}^{M_2}$, the corresponding optimal directions can be utilized to construct a neighborhood set for each point. This is the basis for Direction search Subspace Clustering (DSC) \cite{rahmani2017direction}  -- a new spectral-clustering-based method that uses iPursuit as its core procedure to build a similarity matrix. DSC obtains all the optimal directions by solving one $r \times M_2$ -  dimensional (or $M_2 \times M_2$ if (\ref{eq: sparse rep}) is used) optimization problem. Subsequently, the similarity matrix is formed, to which spectral clustering is applied to cluster the data. We present some results using DSC in Section \ref{sec:dsc} and refer the reader to \cite{rahmani2017direction} for further details.  

\section{Numerical Simulations}
In this section, we present some numerical experiments to study the performance of the proposed subspace clustering algorithm (iPursuit) and compare its performance to existing approaches. 
First, we present some numerical simulations confirming the intuition gained through performance analysis. Then, we compare the run time and performance of Algorithm 1 with existing algorithms to investigate the speed and capability of iPursuit in dealing with non-independent subspaces and noisy data. Subsequently, we apply iPursuit to real data for motion segmentation. \textcolor{black}{ Finally, we present a set of experiments with synthetic and real data (face images) to highlight how the integration of proposed direction search with spectral clustering  can overcome the limitations of iPursuit discussed in Section \ref{sec:restrcition}.}
In the presented experiments, we consider subspaces with intersection. The data in all experiments (except for those with real data) is generated as follows. The given data points lie in a union of $N$ $d$-dimensional subspaces $\{ \calS_i \}_{i = 1}^{N}$.
Define $\calM$ as a random $y$-dimensional subspace. We generate each subspace $\calS_i$ as
$
\calS_i = \calM \oplus \calR_i \: ,
$
where $\calR_i$ is a random $d - y$ dimensional subspace. Thus, the dimension of the intersection of the subspaces $\{ \calS_i \}_{i = 1}^{N}$ is equal to $y$ whp. 
In all simulations, iPursuit refers to Algorithm 1 with the error correction techniques presented in Algorithm 2 and Algorithm 3. 
 With noisy data, we solve (\ref{eq:robust2}) with the 2 data points closest to the least singular vector to choose a final vector $\bq$, whereas with motion segmentation data we use 5 neighboring data points.
In all experiments using synthetic data expect the one in Section \ref{sec:simul_coh_data}, the data points are distributed uniformly at random within the subspaces, i.e., a data point lying in an $r_i$-dimensional subspace $\calS_i$ is generated as $\bV_i \bg$, where the elements of $\bg \in \mathbb{R}^{r_i}$ are sampled independently from a standard normal distribution $\calN(0,1)$. All simulations were performed on a desktop PC with an Intel 3.4 GHz Core i7 processor 
and 8 GB RAM.

\begin{table}
\centering
\caption{Run time of different algorithms ($M_1 = 50 $, $N=3$, $\{r_i \}_{i=1}^3 = 10$, $\{n_i \}_{i=1}^3 = M_2/3$)}
\begin{tabular}{|c|c|c|c|c|c|c| }
\hline
  $M_2$ & iPursuit & SSC  & LRR   & SSC-OMP & TSC & $K$-flats \\
\hline
  300 & 0.14 s & 1.1  s &  0.9 s   & 1.6 s & 0.8 s &  0.1 s \\
\hline
  3000 & 0.35 s & 209 s  &  26 s   & 56 s & 18.5 s  & 0.2 s \\
\hline
  15000 &    2.78 s & $>$ 2 h  & 2800 s   & 5340 s & 4100 s &  1.3 s \\
\hline
  30000 &   10.6 s  & $>$ 2 h  &  $>$ 2 h  & $>$ 2 h  & $>$ 2 h & 2.5 s \\
\hline
\end{tabular}
\label{tab:runnng_M2}
\end{table}

\subsection{The importance of the coherency parameter}
\label{sec:sim_coh}
\begin{figure}
 \centering
    \includegraphics[width=0.5 \textwidth]{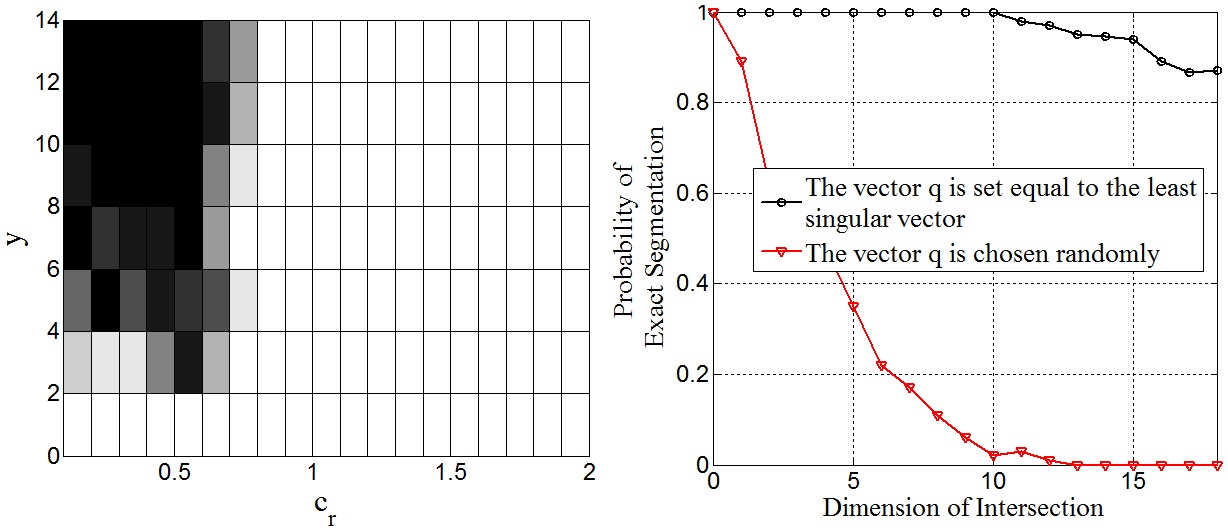}
    \vspace{-0.6cm}
    \caption{Left: Phase transition for various coherency parameters and dimension of the intersection. White designates exact subspace identification. Right: The performance of iPursuit (probability of exact segmentation) versus the dimension of the intersection.}
    \label{mcr}
\end{figure}

In this simulation, it is shown that the performance of iPursuit is improved when $\bq$ is coherent with the innovation subspace. It is assumed that the data lies in two subspaces and $M_1  = 50$. The dimension of the subspaces is equal to 15 and the dimension of their intersection varies between 0 to 14.   Each subspace has 100 data points (total of 200 data points).
The left plot of Fig. \ref{mcr} shows the phase transition in the plane of $c_r$ (the coherency of $\bq$ with the innovation subspace) defined as
$
c_r =  \frac{ \| \bq^T \bP_2 \| }{\| \bq^T \bV_1 \| }
$
and $y$, where $y$ is the dimension of the intersection of the subspaces.
(IP) is used for subspace identification and define $\hat{\bV}_1$ and $\hat{\bV}_2$ as orthonormal bases for the identified subspaces. A trial is considered successful if
\begin{eqnarray}
\| (\bI - \bV_1 \bV_1^T) \hat{\bV}_1 \|_F + \| (\bI - \bV_2 \bV_2^T) \hat{\bV}_2 \|_F \leq 10^{-3} \: .
\label{eq:trial_success}
\end{eqnarray}
As shown, the performance improves as $c_r$ increases. The left plot of Fig. \ref{mcr} shows that when $c_r$ is large enough, iPursuit yields exact segmentation even when $y = 14$. This simulation confirms our analysis regarding the importance of the coherency parameter. 

\subsection{Choosing the vector q}
\label{sec:sim_choose_q}
Next, it is shown that choosing the vector $\bq$ using the technique proposed in Section \ref{sec:choose_q} can substantially improve the performance of the algorithm. In this experiment, the data points are assumed to lie in two 20-dimensional subspaces $\calS_1$ and $\calS_2$ and $M_1 = 50$.
The right plot of Fig. \ref{mcr} shows the probability of correct subspace identification versus the dimension of the intersection. Each point in the plot is obtained by averaging over 100 independent runs and (IP) is used for subspace segmentation. Again, a trial is considered successful if (\ref{eq:trial_success}) is satisfied.
It is clear that when $\bq$ is equal to the least dominant singular vector of the data, the algorithm performs substantially better, i.e., the algorithm is more robust to the intersection between the subspaces. This is because the least dominant singular vector is coherent with the innovation subspace when the subspaces are close to each other.

\subsection{The ratio $\frac{n_1}{n_2}$}
In Lemma \ref{lemma_random}, it was shown that the optimal point of (\ref{convex}) is more likely to lie in the innovation subspace $\mathcal{I}  \left( \mathcal{S}_2 \perp  \mathcal{S}_1 \right)$  if the ratio $\frac{n_1}{n_2}$ increases. In this section, we confirm this analysis numerically. According to the presented analysis and the numerical simulations in Sections \ref{sec:sim_coh}, \ref{sec:sim_choose_q} and \ref{sec:sim_multi}, the algorithm performs very well even when $n_1 = n_2$.
However, to observe the effect of this ratio, we consider a particularly hard subspace clustering scenario with significant intersection between the subspaces. Specifically, we assume that the given data points lie in two 40-dimensional subspaces ($\calS_1$ and $\calS_2$), the dimension of the intersection is equal to 39 and $M_1 = 200$.

\begin{figure}[t!]
 \centering
    \includegraphics[width=0.5\textwidth]{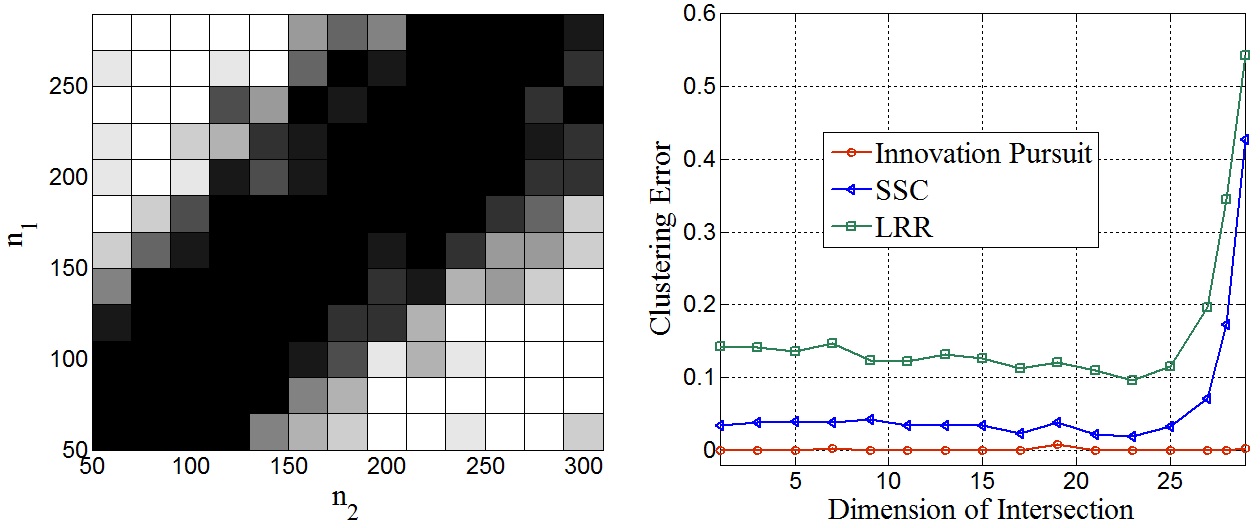}
    \vspace{-0.6cm}
    \caption{Left: Phase transition for different values of $n_1$ and $n_2$, the number of data points in the first and second subspaces. White designates exact subspace identification. Right: The clustering error of iPursuit, SSC and LRR versus the dimension of the intersection.}
    \label{n2n1fig}
\end{figure}

The left plot of Fig. \ref{n2n1fig} shows the phase transition of iPursuit in the plane of $n_1$ and $n_2$  
and (IP) is used for subspace identification. For each $(n_1 , n_2)$, we generate 10 random realizations of the problem. A trial is considered successful if (\ref{eq:trial_success}) is satisfied.
%
Clearly, iPursuit achieves better performance away from the diagonal, i.e., when the ratio $\frac{n_1}{n_2}$ is away from 1. When $\frac{n_1}{n_2}$ is smaller than one, it is more likely that
\begin{eqnarray}
\underset{\bc \in \mathcal{I}  \left( \mathcal{S}_1 \perp  \mathcal{S}_2 \right) \atop \bc^T \bq = 1}{\inf} \: \:  \| \bc^T \bD_1 \|_1 <  \underset{\bc \in \mathcal{I} \left( \mathcal{S}_2 \perp  \mathcal{S}_1 \right) \atop \bc^T \bq = 1}{\inf} \: \:  \| \bc^T \bD_2 \|_1.
\end{eqnarray}
Thus, according to Lemma \ref{lemma_random}, if $\frac{n_2}{n_1}$ increases, the optimal point of (\ref{convex}) is likely to lie in $\mathcal{I}  \left( \mathcal{S}_1 \perp  \mathcal{S}_2 \right)$, and the algorithm yields correct segmentation.

\subsection{Run time comparison}
\label{sec:running}
Solving the proposed direction search optimization problem has low computation complexity. This feature makes the proposed iterative method notably faster than the state-of-the-art clustering algorithms. In this section, we study the run time of the proposed method, SSC \cite{elhamifar2013sparse}, SSC-OMP \cite{dyer2013greedy}, LRR \cite{liu2013robust}, TSC \cite{heckel2013robust}, and $K$-flats \cite{vidal2011subspace}. For SSC and LRR, we use the ADMM solvers provided by the authors, and for TSC we use the code provided by the author.  
For SSC-OMP, the number of neighborhing data points found by the OMP function is set equal to $\min (r/N , 50)$ and for TSC, the value for the number of parameters is set equal to $\min (2 r/N , 50)$. In this simulation, since the generated subspaces are independent, all methods except for the $K$-flats algorithm yield exact subspace clustering for all the scenarios.

\smallbreak
\noindent
\textbf{Table \ref{tab:runnng_M2}:} This table compares the run time of the algorithms for a different number of data points. The data points lie in a union of three 10-dimensional subspaces and $\{n_i\}_{i=1}^3 = M_2/3$. The proposed approach exhibits notable speedup over spectral clustering based methods. The complexity of the $K$-flats algorithm is roughly $\calO( (\max_i r_i) \: r M_2)$ and its speed is comparable to iPursuit, albeit its performance is sensitive to its random initialization. Also, $K$-flats requires prior knowledge of the dimensions and number of subspaces, and its performance  degrades when the subspaces are close.

\smallbreak
\noindent
\textbf{Table \ref{tab:running_N}:} This table studies the run time versus the number of subspaces. The data lies in a union of $N$   $\frac{100}{N}$-dimensional subspaces, $M_1 = 110$, and $M_2 = 5000$. As shown, the run time of iPursuit increases linearly with the number of subspaces. The run time of LRR, SSC, and TSC \textcolor{black}{does not} exhibit strong dependence on the number of subspaces. In our simulation, the run time of SSC-OMP is a decreasing function of $N$ since $r_i = r/N$. Thus, the dimension of the subspaces is larger for smaller $N$ and the OMP function needs to find more data points for the neighborhood of each data point.



\begin{table}
\centering
\caption{Run time of the different algorithms ($r = 100, M_1 = 110, M_2 = 5000$, $\{r_i \}_{i=1}^N = 100/N$, $\{n_i \}_{i=1}^N = M_2/N$) }
\begin{tabular}{|c |c  |c| c | c|c|c| }
\hline
  $N$ & iPursuit & SSC  & LRR    & SSC-OMP & TSC & $K$-flats \\
\hline
  2 & 0.81 s & 756 s   &   120 s  & 352 s & 55 s & 0.26 s  \\
\hline
  5  & 1.41 s &  769 s  &    126 s  & 203 s & 52 s & 0.45 s  \\
\hline
  10 & 2.36 s &  754 s &   136 s   & 164 s & 58 s &  0.71 s \\
\hline
  25 &  5.51 s & 764 s   &  125 s   & 167 s &60 s &  5.21 s \\
\hline
\end{tabular}
\label{tab:running_N}
\end{table}


\subsection{Clustering data in union of multiple subspaces}
\label{sec:sim_multi}
Now we consider a setting where the data points lie in a union of 15 30-dimensional subspaces $\{ \calS_i \}_{i = 1}^{15}$ and $M_1 = 500$. There are 90 data points in each subspace and the distribution of the data in the subspaces is uniformly random.
 In this experiment, we compare the performance of iPursuit to the state-of-the-art SSC \cite{elhamifar2013sparse} and LRR \cite{liu2013robust} algorithms. The number of replicates used in the spectral clustering for SSC and LRR is equal to 20.
Define the clustering error (CE) as the ratio of misclassified points to the total number of data points. The right plot of Fig. \ref{n2n1fig} shows CE versus the dimension of the intersection. The dimension of intersection varies between 1 to 29. Thus, the rank of the data ranges from 436 to 44.
Each point in the plot is obtained by averaging over 40 independent runs. iPursuit is shown to yield the best performance. The proposed algorithm finds the subspaces consecutively, thus all the subspaces are identified in 14 steps.


\subsection{Noisy data}
\label{sec:simul_noise}
\begin{figure}
 \centering
    \includegraphics[width=0.50\textwidth]{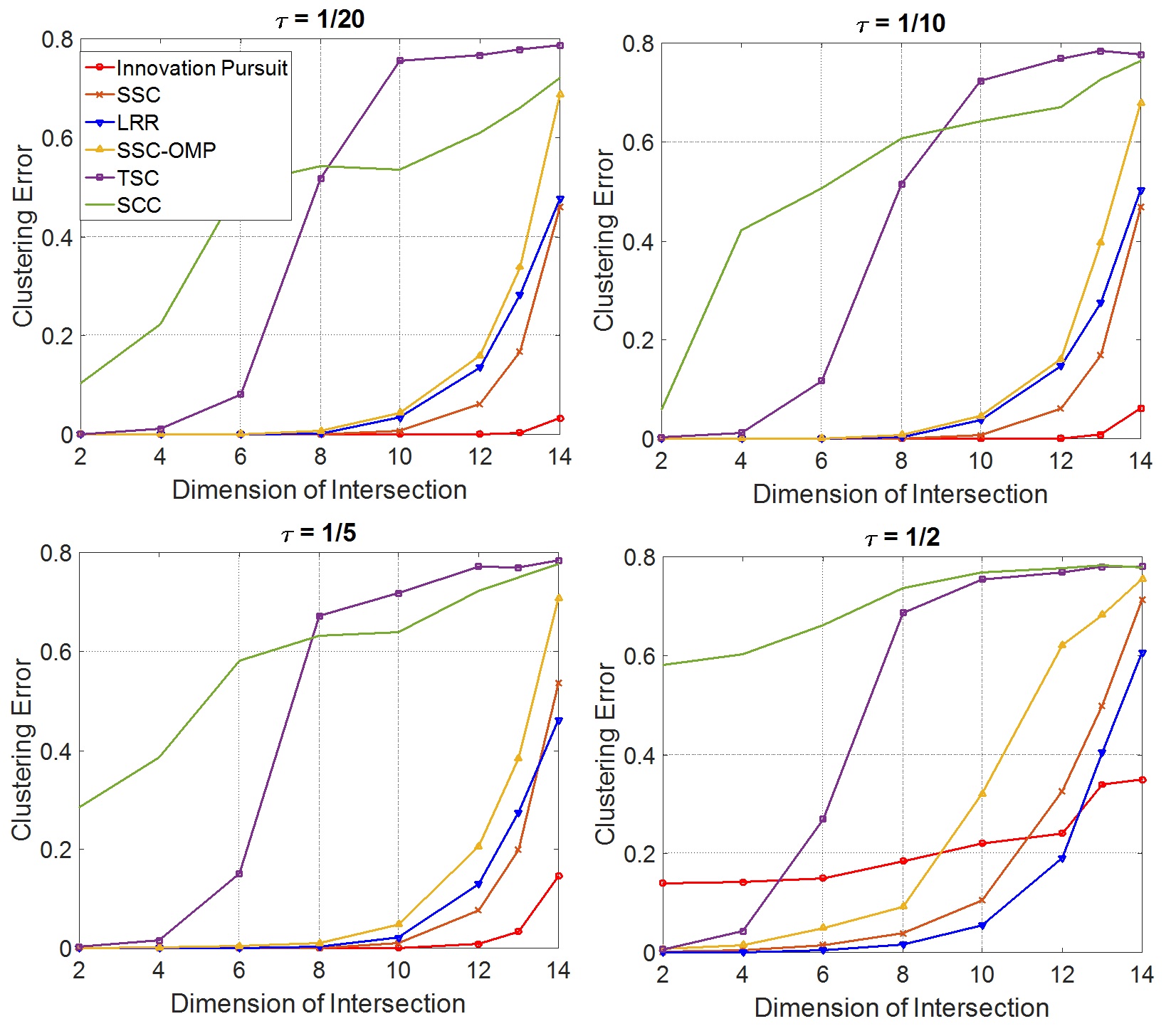}
    \vspace{-0.65cm}
    \caption{Performance of the algorithms versus the dimension of intersection for different noise levels.}
    \label{fig:noise}
\end{figure}

In this section, we study the performance of iPursuit, SSC, LRR, SCC \cite{chen2009spectral}, TSC and SSC-OMP with different noise levels, and varying dimensions of the intersection between the subspaces, which gives rise to both low rank and high rank data matrices. It is assumed that $\bD$ follows Data model 1 with $M_1 = 100$, $M_2 = 500$, $N=6$ and $\{r_i\}_{i = 1}^{6} = 15$. The dimension of the intersection between the subspaces varies from 0 to 14. Thus, the rank of $\bD$ ranges from 20 to 90. The Noisy data follows (\ref{eq:with noise}) and the elements of $\bE$ are sampled independently from a zero mean Gaussian distribution.
Fig. \ref{fig:noise} shows the performance of the different algorithms versus the dimension of the intersection for
$
\tau = \frac{\| \bE \|_F}{\| \bD \|_F}
$
equal to $1/20$, $1/10$, $1/5$ and $1/2$. One can observe that even with $\tau = 1/5$, iPursuit significantly outperforms the other algorithms. In addition, when the data is very noisy, i.e., $\tau = 1/2$, it yields better performance when the dimension of the intersection is large. SSC, LRR, and SSC-OMP yield a better performance for lower dimension of intersection. This is explained by the fact that the rank of the data is high when the dimension of the intersection is low, and the subspace projection operation $\bQ^T \bD_e$ may not always filter out the additive noise effectively.

\begin{table}
\centering
\caption{Clustering error of innovation pursuit with coherent data}
\begin{tabular}{|c  |c| c | c | c|}
\hline
  $\omega$   &  10 & 2 & 0.5 & 0.25  \\
  \hline
  Clustering error ($\%$)       &  0.32 & 0.34 & 0.02 & 12   \\
  \hline
\end{tabular}
\label{tab:coherent_data}
\end{table}

\begin{table*}
\centering
\caption{CE ($\%$) of algorithms on Hopkins155 dataset (Mean - Median).}
\begin{tabular}{|c |c  |c| c | c|c|c| c |}
\hline
  $N$ & iPursuit & SSC  & LRR   & SSC-OMP & TSC & $K$-flats & SCC \\
\hline
  $N=2$ & 3.33 - 0.27 & 1.52 - 0  &  2.13 - 0   & 16.92 - 12.77   & 18.44 - 16.92 &  13.62 - 10.65 & 2.06 - 0  \\
\hline
  $N=3$ & 6.91 - 2.44 & 4.40 - 1.56   &  4.03 - 1.43   & 27.96 - 30.98 & 28.58  - 29.67 &  14.07 - 14.18 &  6.37 - 0.21\\
\hline
\end{tabular}
\label{tab:realdata}
\end{table*}

\subsection{\textcolor{black}{Coherent data points}}
\label{sec:simul_coh_data}
Theorem \ref{two_ind} and Theorem \ref{corol1} established sufficient conditions for iPursuit to yield exact clustering under uniform data distribution in worst case scenarios. In this section, we provide examples with coherent data indicating that these conditions are by no means necessary. 
Similar to the experiment in Section \ref{sec:simul_noise}, the data points lie in a union of 6 15-dimensional subspaces. The dimension of the intersection between the subspaces is equal to 13 and $\tau = 1/5$. However, unlike the experiment in Section  \ref{sec:simul_noise}, here the data points are not distributed uniformly at random within the subspaces. 
A data point in the $i^{\text{th}}$ subspace is generated as $\bV_i \bg $ where $\bg = \ba_i + \omega \hat{\bg}$. The vector $\ba_i$ is a fixed unit vector and $\hat{\bg}$ is sampled uniformly at random from the unit $\ell_2$-norm sphere. Thus, the data points in the $i^{th}$ subspace are concentrated around $\ba_i$ with the coefficient $\omega$ determining how concentrated they are -- a smaller $\omega$ implies the data points are more mutually coherent. Table \ref{tab:coherent_data} provides the clustering error for different values of $\omega$. As shown, decreasing $\omega$ (increasing the coherency between the data points) can even result in improved performance. The reason is two-fold: as the data points become more coherent, the elements of $\bh_1$ corresponding to the subspace in which the constraint vector lies increases, and also the performance of the error correction technique presented in Algorithm 2 improves.
However, this trend does not continue as the data points become more highly concentrated around a given direction (at $\omega = 0.25$, the clustering error increases to $12\%$). In this case, the algorithm cannot obtain an accurate basis for the subspaces due to the rapid decay of the singular values (as the data is highly coherent), the fact that the data is noisy and that the dimension of the intersection is fairly large.


\subsection{Real data}
In this section, we apply iPursuit to the problem of motion segmentation using the Hopkins155 \cite{tron2007benchmark} dataset, which contains video sequences of 2 or 3 motions. The data is generated by extracting and
tracking a set of feature points through the frames \cite{tron2007benchmark}. Most of the videos are checkerboard and traffic videos.
In motion segmentation, each motion corresponds to one subspace. Thus, the problem here is to cluster data lying in two or three subspaces.  Table \ref{tab:realdata} shows CE (in percentage) for iPursuit, SSC, LRR, TSC, SSC-OMP and $K$-flats. We use the results reported in \cite{elhamifar2013sparse,heckel2013robust,vidal2011subspace,park2014greedy}.
\textcolor{black}{
For SSC-OMP and TSC, the  number of parameters for motion segmentation are equal to 8 and 10, respectively}.
One can observe that iPursuit yields competitive results comparable to SSC, SCC, and LRR and outperforms TSC, SSC-OMP and $K$-flats.  



\begin{table}
\centering
\caption{Clustering error of the algorithms ( $M_2 = 1200$, $N=20$, $\{r_i \}_{i=1}^N = 6$, $\{n_i \}_{i=1}^N = M_2/6$) }
\begin{tabular}{|c  |c| c | c|c|c| }
\hline
  $M_1$   & iPursuit+Spectral-Clustering  & SSC  & SSC-OMP& TSC \\
       & (DSC \cite{rahmani2017direction})  &   & &

        \\
\hline
  50   & 0.27    &   3.72
 &  6.23 &   10.68   \\
\hline
  30   & 0.73  &  4.27 &   14.03 &    63.41   \\
\hline
  20   & 2.83  &   15.72 &  29.39 &   71.54   \\
\hline
\end{tabular}
\label{tab:res_1}
\end{table}

\subsection{Innovation pursuit with spectral clustering}
\label{sec:dsc}
In this section, it is shown that integrating iPursuit with spectral clustering can effectively overcome the limitations discussed in Section \ref{sec:restrcition}. For a detailed performance analysis of the integration of innovation pursuit with spectral clustering (the DSC algorithm), we refer the reader to \cite{rahmani2017direction}.
 
1) \textit{Synthetic data:}
Consider data points lying in the union of 20 random 6-dimensional subspaces, each with 60 data points, and the dimension of the intersection between the subspaces is equal to 4. 
When $M_1  > 44$, each subspace has an innovation subspace with dimension 2 w.r.t. the other subspaces whp. 
If $M_1 < 44$, the innovation requirement is not satisfied. Table \ref{tab:res_1} shows CE of various algorithms for different values of $M_1$. The integration of iPursuit with spectral clustering yields accurate clustering even for $M_1 < 44$ and outperforms other spectral clustering based methods even when the subspaces lack relative innovation. 
\begin{figure}
 \centering
    \includegraphics[width=0.50\textwidth]{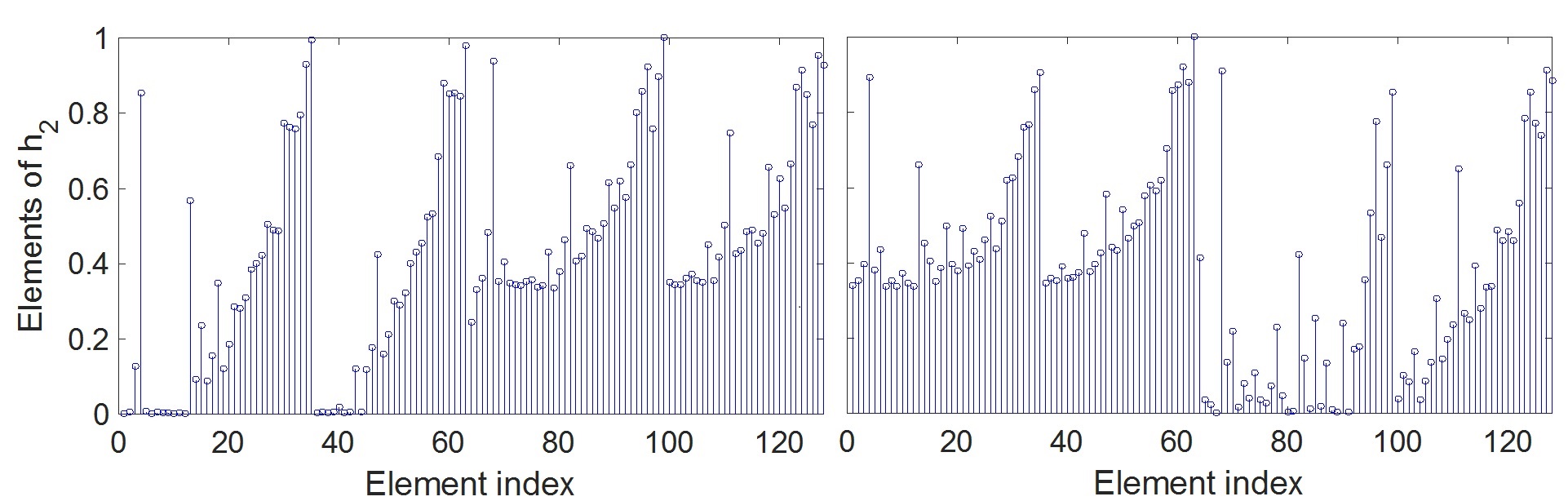}
    \vspace{-0.25cm}
    \caption{The data is formed with 128 face images of two individuals. {Left}: The elements of vector $\bh_2$, where the first column is used as the constraint vector. {Right}: The elements of vector $\bh_2$, where the $80^{\text{th}}$ column is used as the constraint vector.}
    \label{fig:faces_projec}
\end{figure}
%

2) \textit{Face clustering:}
Face clustering is an expressive real world example to study the second limitation discussed in Section \ref{sec:restrcition}. We use the Extended Yale B dataset, which contains 64 images for each of 38 individuals in frontal
view and different illumination conditions \cite{lee2005acquiring}. The faces for each subject can be approximated with a low-dimensional subspace. Thus, a data set containing face images from multiple subjects can be modeled as a union of subspaces. 

For illustration, we compose a data matrix as $\bD = [\bD_1 \: \: \bD_2]$, where $\bD_1 \in \mathbb{R}^{M_1 \times 64}$ and $\bD_2 \in \mathbb{R}^{M_1 \times 64}$ contain the vectorized images of two individuals (data consists of two clusters). We solve (\ref{eq: sparse rep}) and use the first column as the constraint vector.  We form the orthonormal basis $\bF_1$ as described in Step 4 of Algorithm 1. The columns of $\bF_1$ are supposed to span the column space of $\bD_1$. 
The left plot of Fig. \ref{fig:faces_projec} shows the $\ell_2$-norm of the columns of $(\bI - \bF_1 \bF_1^T)\bD$. One can observe that a notable part of the columns of the first cluster have remarkable projections on the complement space of $\text{span}(\bF_1)$.  
A similar behavior is observed if we choose a data point from the second cluster as a constraint vector as for the right plot of Fig. \ref{fig:faces_projec} 
where the $80^{\text{th}}$ column is used. 
Thus, we cannot obtain a proper basis for the clusters. This is due to the fact that the data points within each cluster approximately form sub-clusters with relative innovation. 

We apply DSC, the integration of iPursuit with spectral clustering, to face clustering and present results for a different number of clusters in Table \ref{tab:faces_result}.
The performance is also compared with SSC, SCC, and TSC. Heretofore, SSC yielded the best known result for this problem. 
For each number of clusters shown (except 38), we ran the algorithms over 50 different random combinations of subjects from the 38 clusters. To expedite the runtime, we project the data on the span of the first 500 left singular vectors, which does not affect the performance of the algorithms (expect SSC). 
For SSC, we report the results without projection (SSC) and with projection (SSC-P). As shown, DSC yields accurate clustering and notably outperforms the performance achieved by SSC.
\begin{table}
\centering
\caption{Clustering error ($\%$) of different algorithms on the Extended Yale B dataset.  }
\begin{tabular}{|c  |c| c | c|c|c| }
\hline
   $\#$ of    & iPursuit +   &   &   &  &  \\
    subjects    & spectral clustering   &  SSC &  SSC-P & SCC &  TSC \\
       & (DSC \cite{rahmani2017direction})   &   &   &  &  \\
\hline
  5  & \textbf{ 2.56  } & 4.24  & 29.04  & 62.62 & 25.62 \\
\hline
  10  & \textbf{4.88 }   &  9.53 & 32.76 & 74.13  & 40.46 \\
\hline
15  & \textbf{4.71}   & 15.66   & 34.21  & 77.02  & 44.79 \\
\hline
 20   & \textbf{6.45}   &  19.95  & 33.67 & 78.50 & 45.30 \\
\hline
 25   & \textbf{8.53}   & 24.76
  & 50.19 & 79.37 & 46.46 \\
  \hline
   38   & \textbf{8.84}   & 27.47
  & 50.37 & 88.86 & 47.12 \\
\hline
\end{tabular}
\label{tab:faces_result}
\end{table}

%
%
\section*{Appendix}
\noindent
\textbf{Simplifying the requirements of Theorem \ref{corol1}} \\
According to Theorem \ref{two_ind}, when the data points are well distributed within the subspaces and there is a sufficient number of data points in the subspaces (say $\calS_1$), the optimal point of (\ref{convex}) lies in $\calI (\calS_2 \perp \calS_1)$ whp even if $\bq$ is coherent with $\calS_1$.
This intuition is the basis for an unproven conjecture we use herein to simplify the sufficient conditions of Theorem \ref{corol1}.
We conjecture that the optimal point of the non-convex optimization problem (\ref{nime_convex}) lies in one of the innovation subspaces if there are enough data points in the subspaces and they are well distributed. First, to avoid cumbersome notation let
$
\calI_{j} := \calI \left( \calS_{j} \perp \overset{n}{\underset{k = 1 \atop k \neq {j} }{\oplus}} \calS_k \right).
$
Then, the conjecture is tantamount to saying that, if the data follows Data model 1 with $n$ subspaces, then it is highly likely that
\begin{eqnarray}
\begin{aligned}
\underset{\delta \in \overset{n}{\underset{k = 1 }{\oplus}} \calS_k  \atop \| \delta \| = 1}{\inf} \sum_{\bd_i \in \overset{n}{\underset{k = 1 }{\cup}} \bD_k} \left| \mathbf{\delta}^T \bd_i \right|
& = \underset{j}{\min} \left( \underset{\delta \in \calI_j \atop \| \delta \| = 1}{\inf} \sum_{\bd_i \in  \bD_j} \left| \mathbf{\delta}^T \bd_i \right|  \right) \\
& = \underset{\delta \in \calI_{t_n} \atop \| \delta \| = 1}{\inf} \sum_{\bd_i \in  \bD_{t_n}} \left| \mathbf{\delta}^T \bd_i \right|  \: ,
\end{aligned}
\label{conjbas}
\end{eqnarray}
%
for some integer $ 1\leq t_n \leq n$. 
Adopting the same notation preceding Theorem \ref{corol1}, we redefine $\calI_{j}$ as
 $
\calI_{j} := \calI \left( \calS_{j} \perp \overset{m}{\underset{k = 1 \atop k \neq {j} }{\oplus}} \calS_k \right).
$
Accordingly, the sufficient condition provided in Theorem \ref{corol1} can be simplified as
\begin{eqnarray}
\begin{aligned}
& \frac{1}{2}  \underset{\delta \in \calI_{t_{m-1}}  \atop \| \delta \| = 1}{\inf} \sum_{\bd_i \in  \bD_{t_{m-1}}} \hspace{-4mm} \left| \mathbf{\delta}^T \bd_i \right|
\hspace{-0.5mm}  
> \| \bV_m^T  \bT_{m-1} \| \bigg( \|\alpha_m \| +  n_{0m} \bigg),\\
\\
& \frac{ \| \bq_m^T \bP_m \|}{ 2\| \bq_m^T \bT_{m-1} \|}  \left( \underset{\delta \in \calI_{t_{m-1}} \atop \| \delta \| = 1}{\inf} \sum_{\bd_i \in  \bD_{t_{m-1}}} \left| \mathbf{\delta}^T \bd_i \right| \right)   \\
& \qquad\qquad\qquad \qquad\qquad
>  \| \bV_m^T \bP_m \| \bigg( \| \alpha_m \| + n_{0m} \bigg),
\end{aligned}
\label{comlxinq}
\end{eqnarray}
%
where $t_{m-1}$ is defined similar to $t_n$ in (\ref{conjbas}). Define $\bD_{t_{m-1}}^i$ as the projection of the columns of $\bD_{t_{m-1}}$ on $\calI_{t_{m-1}}$.
The infimum on the LHS of (\ref{comlxinq}) is the permeance static for the columns of $\bD_{t_{m-1}}^i$ in $\calI_{t_{m-1}}$. According to conditions (\ref{comlxinq}), it is evident that three factors are important for disjoining $\calS_m$ and $\{ \calS_i \}_{i=1}^{m-1}$, namely, \\
1. The distribution of the data points in the subspaces $\{ \calS_i \}_{i=1}^{m-1}$ (especially the columns of $\bD_{t_{m-1}}^i$) since it determines the value of the permance statistic.\\
2. The coherency of $\bq_m$ with  $\mathcal{I} \big( \mathcal{S}_m \perp \overset{m-1}{\underset{k = 1 }{\oplus}} \calS_k \big)$. \\
3. The distances between the subspaces $\{ \calS_i \}_{i=1}^{m-1}$: if the subspaces are too close to each other, the columns of $\bD_{t_{m-1}}$ will have small projections on $\calI_{t_{m-1}}$. 

\smallbreak
\noindent
\noindent\textbf{Proof of Lemma \ref{new_lemma}} \\
Assume the conditions of Lemma \ref{new_lemma} are satisfied and define the vector $\bg$ as the projection of $\bc^{*}$ onto $\calS_2$.
For contradiction, assume that the columns of $\bD_2$ corresponding to the non-zero elements of $\bg^T \bD_2$ do not span $\calS_2$. Let $\bD_2^o$ and $\bD_2^p$ denote the columns of $\bD_2$ orthogonal to ${\bg}$ 
and not orthogonal to ${\bg}$, respectively.
In addition, define $\calS_2^o$ as the span of $\bD_2^o$  and $\calS_2^p$ as the span of $\bD_2^p$. The subspaces $\calS_2^o $ and $\calS_2^p $ lie within $\calS_2$ and their individual dimension is less than $r_2$.
The subspace $\calS_2^p$ is not a subset of $\calS_2^o$ because ${\bg}$ is orthogonal to $\calS_2^o$ but not to $\calS_2^p$. Also, $\calS_2^o$ cannot be a subset of $\calS_2^p$ because the dimension of  $\calS_2^p$ is less than $r_2$. Hence, $\bD_2$ can follow Data model 1 with $N \ge 2$, which leads to a contradiction. \\
\\
\textbf{Proof of Theorem \ref{two_ind}} \\
The main idea is to show that $\bc_2$ is the optimal point of (\ref{convex}). Thus, we want to show that
\begin{eqnarray}
\underset{\bc \in  \left( \mathcal{S}_2 \oplus \mathcal{S}_1 \right) \atop \bq^T\bc = 1}{\arg\min} \: \:  \| \bc^T \bD \|_1 = \underset{\bc \in \mathcal{I} \left( \mathcal{S}_2 \perp \mathcal{S}_1 \right) \atop \bq^T \bc = 1}{\arg\min} \: \:  \| \bc^T \bD_2 \|_1
\end{eqnarray}
Define $g(\delta) $ as
\begin{eqnarray}
g(\delta)  =  \| (\bc_2 - \mathbf{\delta})^T \bD \|_1 - \| \bc_2^T \bD \|_1  \: .
\label{g_delta}
\end{eqnarray}
Since (\ref{convex}) is convex, it suffices to check that $g(\delta) >0$ for every sufficiently small non-zero perturbation $\delta$ such that
\begin{eqnarray}
\delta^T \bq = 0 \: \: , \: \: \delta \in \mathcal{S}_1 \oplus \mathcal{S}_2 \: .
\label{condition_asli}
\end{eqnarray}
The conditions on $\delta$ are to ensure that $\bc_2 - \mathbf{\delta}$ is a feasible point of (\ref{convex}).
If $\bc_2$ is the optimal point of (OP) in (\ref{eq:oracel}), then the cost function is increased when we move from the optimal point along a feasible perturbation direction. Observe that $\bc_2 - \delta_2$ is a
feasible point of (\ref{eq:oracel}) if and only if the perturbation $\delta_2$ satisfies
\begin{eqnarray}
\mathbf{\delta}_2^T \bq = 0 \: \: , \: \: \mathbf{\delta}_2 \in\mathcal{I} \left( \mathcal{S}_2 \perp  \mathcal{S}_1 \right).
\label{condition_oracel}
\end{eqnarray}
Therefore, for any non-zero $\mathbf{\delta}_2$ which satisfies (\ref{condition_oracel})
\begin{eqnarray}
\| (\bc_2 - \mathbf{\delta}_2)^T \bD_2 \|_1 - \| \bc_2^T \bD_2 \|_1 > 0 \: .
\label{strict_oracel}
\end{eqnarray}
When $\delta_2 \to 0$, we can rewrite (\ref{strict_oracel}) as
\begin{align}
& \| (\bc_2 - \mathbf{\delta}_2)^T \bD_2 \|_1 - \| \bc_2^T \bD_2 \|_1 \nonumber\\
&= \sum_{\bd_i \in \bD_2} \left[ (\bc_2 - \mathbf{\delta}_2)^T \bd_i)^2 \right]^{1/2} - \sum_{\bd_i \in \bD_2} \left| \bc_2^T \bd_i \right| \nonumber \\
& =  \sum_{\bd_i \in \bD_2} \hspace{-.2cm}\left[ (\bc_2^T \bd_i)^2 \hspace{-.08cm} - \hspace{-.08cm}  2 (\bc_2^T \bd_i)(\mathbf{\delta}_2^T \bd_i) \hspace{-.1cm}+\hspace{-.1cm} (\mathbf{\delta}_2^T \bd_i)^2 \right]^{1/2} \hspace{-.15cm} - \hspace{-.2cm} \sum_{\bd_i \in \bD_2} \left| \bc_2^T \bd_i \right| \nonumber\\
&=  \sum_{\bd_i \in \bD_2 \atop i \in \calL_0} \left| \delta_2^T \bd_i \right|
+ \sum_{\bd_i \in \bD_2 \atop i \in \calL^c_0} \left|\bc_2^T \bd_i \right| \bigg[ 1 - 2 \frac{\sgn (\bc_2^T \bd_i)}{|\bc_2^T \bd_i|} (\mathbf{\delta}_2^T \bd_i) \nonumber\\
& \qquad\qquad\qquad\qquad +   \calO (\| \delta_2 \|^2 )\bigg]^{1/2}  - \sum_{\bd_i \in \bD_2 \atop i \in \calL^c_0} \left| \bc_2^T \bd_i \right| \nonumber\\
&= \sum_{\bd_i \in \bD_2 \atop i \in \calL_0} \left| \delta_2^T \bd_i \right|- \hspace{-0.25cm}\sum_{\bd_i \in \bD_2 \atop i \in \calL^c_0} {\sgn} (\bc_2^T \bd_i) (\mathbf{\delta}_2^T \bd_i) + \mathcal{O} ( \|\mathbf{\delta_2}\|^2)
\end{align}
where the last identity follows from the Taylor expansion of the square root. Thus,
\begin{eqnarray}
\sum_{\bd_i \in \bD_2 \atop i \in \calL_0} \left| \delta_2^T \bd_i \right| - \sum_{\bd_i \in \bD_2 \atop i \in \calL^c_0} {\sgn} (\bc_2^T \bd_i) (\mathbf{\delta}_2^T \bd_i) + \mathcal{O} ( \|\mathbf{\delta}_2\|^2)
\end{eqnarray}
has to be greater than zero for small $\delta_2$ which satisfies (\ref{condition_oracel}). Therefore,
\begin{eqnarray}
\begin{aligned}
&\sum_{\bd_i \in \bD_2 \atop i \in \calL_0} \left| \delta_2^T \bd_i \right| - \sum_{\bd_i \in \bD_2 \atop i \in \calL^c_0} {\sgn} (\bc_2^T \bd_i) (\mathbf{\delta}_2^T \bd_i) \ge 0 \: \: \: \forall \: \delta_2 \in \mathbb{R}^{M_1} \\
& \text{s.t.} \quad \delta_2^T \bq = 0 \: \:, \: \: \delta_2 \in \mathcal{I}(\mathcal{S}_2 \perp \mathcal{S}_1 ).
\end{aligned}
\label{observation}
\end{eqnarray}

To simplify $g(\delta)$, we decompose $\delta$ into
$
\delta = \delta_1 + \delta_I
$
where $\delta_1 \in \mathcal{S}_1$ and $\delta_I \in \mathcal{I} \left( \mathcal{S}_2 \perp \mathcal{S}_1 \right)$. The vectors $\bc_2$ and $\delta_I$ lie in $\mathcal{I}(\mathcal{S}_2 \perp \mathcal{S}_1 )$ which is orthogonal to $\mathcal{S}_1$. Therefore, for the data points in $\mathcal{S}_1$
\begin{eqnarray}
\| (\bc_2 - \mathbf{\delta})^T \bD_1 \|_1 - \| \bc_2^T \bD_1 \|_1  = \sum_{\bd_i \in \bD_1} \left| \mathbf{\delta}_1^T \bd_i \right| \: .
\label{ghesmate_1}
\end{eqnarray}

In addition, as $\delta \to 0$,
\begin{eqnarray}
\begin{aligned}
&\| (\bc_2 - \mathbf{\delta})^T \bD_2 \|_1 - \| \bc_2^T \bD_2 \|_1  =  \\
& \sum_{\bd_i \in \bD_2 \atop i \in \calL_0} \left| \delta^T \bd_i \right| - \sum_{\bd_i \in \bD_2 \atop i \in \calL^c_0} \sgn (\bc_2^T \bd_i) \: \delta^T \bd_i + \mathcal{O} (\| \delta^2 \|) \: .
\end{aligned}
\label{ghesmate_2}
\end{eqnarray}

Therefore, according to (\ref{g_delta}), (\ref{ghesmate_1}) and (\ref{ghesmate_2}), it is enough to show that
\begin{eqnarray}
\begin{aligned}
& g(\delta) = \sum_{\bd_i \in \bD_1} \left| \mathbf{\delta}_1^T \bd_i \right|  + \sum_{\bd_i \in \bD_2 \atop i \in \calL_0} \left| \delta^T \bd_i \right| \\
& \qquad\qquad- \sum_{\bd_i \in \bD_2 \atop i \in \calL^c_0} \sgn (\bc_2^T \bd_i) \: \delta^T \bd_i > 0 \:,
\end{aligned}
\end{eqnarray}
for every $\delta \neq 0$ which satisfies (\ref{condition_asli}). 
According to (\ref{eq:alpha_def}), $g(\delta)$ is equivalent to
\begin{eqnarray}
g(\delta) = \sum_{\bd_i \in \bD_1} \left| \mathbf{\delta}_1^T \bd_i \right| + \sum_{\bd_i \in \bD_2 \atop i \in \calL_0} \left| \delta^T \bd_i \right| -  \delta_1^T \alpha - \delta_I^T  \alpha
\end{eqnarray}
To show that $g(\delta)$ is greater than zero, it suffices to ensure that
\begin{eqnarray}
\begin{aligned}
& \frac{1}{2} \sum_{\bd_i \in \bD_1} \left| \mathbf{\delta}_1^T \bd_i \right| - \sum_{\bd_i \in \bD_2 \atop i \in \calL_0}  \left| \delta_1^T \bd_i \right| > \delta_1^T \alpha \\
& \frac{1}{2} \sum_{\bd_i \in \bD_1} \left| \mathbf{\delta}_1^T \bd_i \right| > \delta_I^T  \alpha -  \sum_{\bd_i \in \bD_2 \atop i \in \calL_0}  \left| \delta_I^T \bd_i \right|
\end{aligned}
\label{cond_lemma}
\end{eqnarray}
For the first inequality of (\ref{cond_lemma}), it is enough to ensure that a lower bound on the LHS is greater than an upper bound on the RHS. Thus, it suffices to have
\begin{eqnarray}
\begin{aligned}
& \frac{1}{2} \underset{\delta \in \mathcal{S}_1  \atop \| \delta \| = 1}{\inf} \sum_{\bd_i \in \bD_1} \left| \mathbf{\delta}^T \bd_i \right|  >  \underset{\delta \in \mathcal{S}_1  \atop \| \delta \| = 1}{\sup} \sum_{\bd_i \in \bD_2 \atop i \in \calL_0}  \left| \delta_1^T \bd_i \right| +    \underset{\delta \in \mathcal{S}_1  \atop \| \delta \| = 1}{\sup} \delta^T \alpha
\end{aligned}
\label{sup_min1}
\end{eqnarray}
The matrices $\bV_1$ and $\bV_2$ were defined as orthonormal bases for $\mathcal{S}_1$ and $\mathcal{S}_2$, respectively. The vector $\alpha$ lies in $\mathcal{S}_2$. Therefore,
\begin{eqnarray}
\begin{aligned}
& \underset{\delta \in \mathcal{S}_1  \atop \| \delta \| = 1}{\sup} \delta^T \alpha =  \| \bV_1^T \bV_2 \| \: \|\alpha\|
\end{aligned}
\end{eqnarray}
In addition, the first term of the RHS of (\ref{sup_min1}) can be bounded by $\| \bV_1^T \bV_2 \| n_0$. Thus, the first inequality of (\ref{lemma_inq}) results from the first inequality of (\ref{cond_lemma}).

Observe that the second inequality of (\ref{cond_lemma}) is homogeneous in $\delta$ since
\begin{eqnarray}
\delta_1^T \bq = - \delta_I^T \bq.
\end{eqnarray}
We scale $\delta$ such that $\delta_1^T \bq = - \delta_I^T \bq = 1$. To ensure that the second inequality of (\ref{cond_lemma}) is satisfied, it is enough to show that
\begin{eqnarray}
\begin{aligned}
&\frac{1}{2} \underset{\delta_1 \in \mathcal{S}_1  \atop \delta_1^T \bq = 1}{\inf} \: \: \sum_{\bd_i \in \bD_1} \left| \mathbf{\delta}_1^T \bd_i \right|  > \\
& \underset{\delta_I \in \mathcal{I} \left( \mathcal{S}_2 \perp \mathcal{S}_1 \right)  \atop \delta_I^T \bq = 1}{\sup} \: \: \left( \delta_I^T  \alpha -  \sum_{\bd_i \in \bD_2 \atop i \in \calL_0}  \left| \delta_I^T \bd_i \right| \right)
\label{linear_cond}
\end{aligned}
\end{eqnarray}
Let us decompose $\delta_I$ into
$
\delta_I = \delta_{ip} + \delta_{iq}
$
where
\begin{eqnarray}
\delta_{ip} = (\bI - \bq^{'} (\bq^{'})^T) \delta_I \: \: , \: \: \delta_{iq} = \bq^{'} (\bq^{'})^T \delta_I
\end{eqnarray}
and $\bq^{'}$ is defined as
$
\bq^{'} = \bP_2 \bP_{2}^T \bq  \:/ \: \| \bP_2 \bP_{2}^T \bq \|
$
where $\bP_2$ was defined as an orthonormal basis for $\mathcal{I} \left( \mathcal{S}_2 \perp \mathcal{S}_1 \right)$.

For the second inequality, it is enough to show that the LHS of (\ref{linear_cond}) is greater than
\begin{eqnarray}
\underset{\delta_I \in \mathcal{I} \left( \mathcal{S}_2 \perp \mathcal{S}_1 \right)  \atop \delta_I^T \bq = 1}{\sup}  \left( \delta_{ip}^T  \alpha + \delta_{iq}^T  \alpha -  \sum_{\bd_i \in \bD_2 \atop i \in \calL_0}  \left| \delta_{ip}^T \bd_i \right|  +  \sum_{\bd_i \in \bD_2 \atop i \in \calL_0}  \left| \delta_{iq}^T \bd_i \right| \right)
\end{eqnarray}
According to the definition of $\bq^{'}$ and $\delta_{ip}$,
\begin{eqnarray}
\delta_{ip} \in  \mathcal{I} \left( \mathcal{S}_2 \perp \mathcal{S}_1 \right) \quad \text{and} \quad \delta_{ip}^T \bq = 0 \: .
\end{eqnarray}
Therefore, according to (\ref{observation}), $\delta_{ip}^T \alpha - \sum_{\bd_i \in \bD_2 \atop i \in \calL_0}  \left| \delta_{ip}^T \bd_i \right| \leq 0 \: .$
Thus, it suffices to show that the LHS of (\ref{linear_cond}) is greater than
\begin{eqnarray}
\begin{aligned}
& \underset{\delta_I \in \mathcal{I} \left( \mathcal{S}_2 \perp \mathcal{S}_1 \right)  \atop \delta_I^T \bq = 1}{\sup}  \: \: \left( \delta_{iq}^T  \alpha +  \sum_{\bd_i \in \bD_2 \atop i \in \calL_0}  \left| \delta_{iq}^T \bd_i \right| \right) \\
& = \frac{1}{\left\| \bq^T \bP_2 \right\|} \left( \left| \alpha^T \bq^{'} \right| + \sum_{\bd_i \in \bD_2 \atop i \in \calL_0}  \left|  \bd_i^T \bq^{'} \right| \right)   \\
&  \leq \frac{ \| \bV_2^T \bP_2 \|}{\left\| \bq^T \bP_2 \right\|} \left( \|\alpha \| + n_0  \right) \:.
\end{aligned}
\label{dovom_1}
\end{eqnarray}
In addition, the LHS of (\ref{linear_cond}) can be simplified as
\begin{eqnarray}
\begin{aligned}
\frac{1}{2} \underset{\delta_1 \in \mathcal{S}_1  \atop \delta_1^T \bq = 1}{\inf} \: \: \sum_{\bd_i \in \bD_1} \left| \mathbf{\delta}_1^T \bd_i \right| \ge \frac{1}{2\| \bq^T\bV_1 \|} \underset{\delta_1 \in \mathcal{S}_1  \atop \| \delta_1 \|= 1}{\inf} \: \: \sum_{\bd_i \in \bD_1} \left| \mathbf{\delta}_1^T \bd_i \right|.
\end{aligned}
\label{dovom_2}
\end{eqnarray}
According to (\ref{linear_cond}), (\ref{dovom_1}) and (\ref{dovom_2}), the second inequality of (\ref{lemma_inq}) guarantees that the second inequality of (\ref{cond_lemma}) is satisfied. Thus, if (\ref{lemma_inq}) is satisfied, then (\ref{cond_lemma}) is satisfied.
\\
\\
\textbf{Proof of Lemma \ref{lemma_random}}\\
To prove Lemma \ref{lemma_random}, we need to find a lower bound on the permeance statistic and an upper bound on $\| \alpha \|_2$. The vector $\alpha$ was defined as
\begin{eqnarray}
\begin{aligned}
&\alpha =  \sum_{\bd_i \in \bD_2} \sgn (\bc_2^T \bd_i) \: \bd_i  =\sum_{ ( \bV_2 \bx_i = \bd_i) \in \bD_2} \sgn (\bc_2^T \bd_i) \: \bV_2 \bx_i
\end{aligned}
\end{eqnarray}
where $\bx_i \in \mathbb{R}^{r_2 \times 1}$ is defined such that $\bV_2 \bx_i = \bd_i$ for $\bd_i \in \bD_2$. In Lemma \ref{lemma_random}, it is assumed that the data points in the subspaces are distributed randomly and the data points are normalized (i.e., the $\ell_2$-norm of the data points is equal to one). Therefore, the vectors $\{ \bx_i \}$ can be modeled as i.i.d. random vectors uniformly distributed on the unit sphere $\mathbb{S}^{r_2 - 1}$ in $\mathbb{R}^{r_2}$. Based on this observation, we make use of the following lemmas from \cite{lerman2015robust} and \cite{foucart2013mathematical} to obtain (\ref{random_suf}).
\begin{lemma}
(Lower bound on the permeance statistic from \cite{lerman2015robust}) Suppose that $\bg_1 , ... , \bg_n$ are i.i.d. random vectors uniformly distributed on the unit sphere $\mathbb{S}^{r - 1}$ in $\mathbb{R}^{r}$. When $r = 1$,
$
\underset{\| \delta \| = 2}{\inf} \: \: \sum_{i = 1}^n \left| \mathbf{\delta}^T \bg_i \right| = 1.
$
When $r \ge 2$, for all $t \ge 0$,
\begin{eqnarray}
\underset{\| \delta \| = 2}{\inf} \: \: \sum_{i = 1}^n \left| \mathbf{\delta}^T \bg_i \right| > \sqrt{\frac{2}{\pi}} \frac{n}{\sqrt{r}} - 2\sqrt{n} -t \sqrt{\frac{n}{r -1 }}
\end{eqnarray}
 with probability at least $1 - \exp(- t^2 / 2) \:$.
\end{lemma}

\begin{lemma}
If $\bg_1 , ... , \bg_n$ are i.i.d. random vectors uniformly distributed  on the unit sphere $\mathbb{S}^{r - 1}$ in $\mathbb{R}^{r}$, then for all $t > 1$
\begin{eqnarray}
\begin{aligned}
& \mathbb{P} \left( \left\| \sum_{i = 1}^n h_i \bg_i    \right\|  \ge \| \bh\| t   \right) \\
& \leq \exp \left( \frac{r}{2}(t^2 - log(t^2) -1) \right) \:.
\end{aligned}
\end{eqnarray}
\end{lemma}

{
\bibliography{IEEEabrv,bibfile}}
\bibliographystyle{IEEEtran}


\end{document}